\documentclass[10pt,twocolumn,letterpaper]{article}

\usepackage[pagenumbers]{cvpr} 

\usepackage[colorlinks=true, linkcolor=blue, citecolor=blue, urlcolor=blue]{hyperref}

\definecolor{cvprblue}{rgb}{0.21,0.49,0.74}

\usepackage[accsupp]{axessibility}  

\definecolor{darkgreen}{rgb}{0,0.5,0}
\definecolor{darkred}{rgb}{0.6,0.0,0.0} 
\newcommand{\best}[1]{\textbf{#1}}
\usepackage{multirow}
\usepackage{xcolor}
\def\paperID{16834} 
\def\confName{CVPR}
\def\confYear{2026}

\title{Multi-view Pyramid Transformer: Look Coarser to See Broader}

\newcommand\CoAuthorMark{\footnotemark[\arabic{footnote}]}

\author{
Gyeongjin Kang$^{1}$\thanks{Equal contribution}\quad
Seungkwon Yang$^{2}$\protect\CoAuthorMark\quad
Seungtae Nam$^{2}$\quad
Younggeun Lee$^{1}$\\
Jungwoo Kim$^{2}$\quad
Eunbyung Park$^{2}$\thanks{Corresponding author}
\vspace{2mm} \\
$^1$Sungkyunkwan University\hspace{2.2mm}$^2$Yonsei University
\vspace{2mm} \\
{\small \url{https://gynjn.github.io/MVP/}}
}

\begin{document}
\twocolumn[{
\renewcommand\twocolumn[1][]{#1}
\maketitle
\begin{center}
    \centering
    \captionsetup{type=figure}
    \vspace{-1.0em}
    \includegraphics[width=1.0\textwidth]{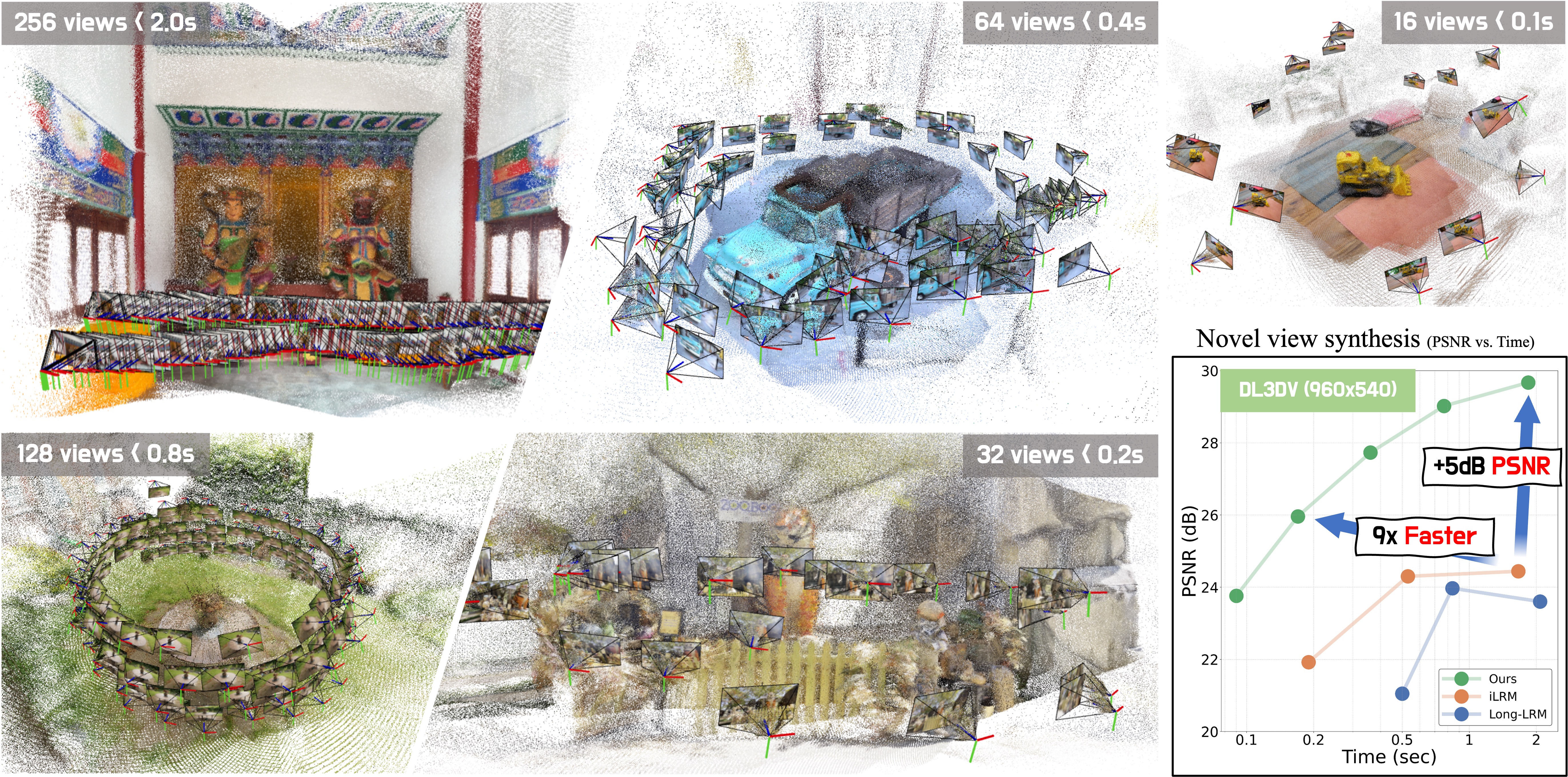}
    \vspace*{-6mm}
    \captionof{figure}{Our method efficiently processes a wide range of input views,
    reconstructing diverse large-scale scenes in under 0.1–2.0 seconds. We utilize Viser~\cite{yi2025viser} for 3D scene visualization. Each marker in the plot represents performance under different numbers of input views: 16, 32, 64, 128, and 256 views for Ours, and 16, 32, and 64 views for both iLRM and Long-LRM. See Tab.~\ref{tab:quantitative result on dl3dv} and~\ref{tab:quantitative result on dl3dv2} for further details.
    }
    \label{fig:teaser}
\end{center}
}]

\renewcommand{\thefootnote}{*}
\footnotetext{Equal contribution}
\renewcommand{\thefootnote}{†}
\footnotetext{Corresponding author}

\begin{abstract}
We propose Multi-view Pyramid Transformer (MVP), a scalable multi-view transformer architecture that directly reconstructs large 3D scenes from tens to hundreds of images in a single forward pass. Drawing on the idea of ``looking broader to see the whole, looking finer to see the details," MVP is built on two core design principles: 1) a local-to-global inter-view hierarchy that gradually broadens the model's perspective from local views to groups and ultimately the full scene, and 2) a fine-to-coarse intra-view hierarchy that starts from detailed spatial representations and progressively aggregates them into compact, information-dense tokens. This dual hierarchy achieves both computational efficiency and representational richness, enabling fast reconstruction of large and complex scenes. We validate MVP on diverse datasets and show that, when coupled with 3D Gaussian Splatting as the 3D representation, it achieves state-of-the-art generalizable reconstruction quality while maintaining high efficiency and scalability across a wide range of view configurations.
\end{abstract}
\vspace{-2mm}
\section{Introduction}
\vspace{-1mm}
\label{sec:intro}

Transformer architectures~\cite{vaswani2017attention} have profoundly reshaped the landscape of modern artificial intelligence. Its self-attention mechanism enables effective modeling of long-range dependencies, leading to significant advances across diverse domains such as natural language processing~\cite{brown2020language, kaplan2020scaling}, computer vision~\cite{dosovitskiy2020vit, caron2021dino}, and multimodal learning~\cite{radford2021learning, kim2024openvla}. This architectural paradigm has further demonstrated remarkable success in 3D vision tasks, including 3D reconstruction~\cite{hong2024lrm, wang2024dust3r, wang2025vggt} and novel view synthesis~\cite{flynn2024quark, jin2025lvsm}.

Recent methods utilizing transformer-based architectures for 3D vision tasks~\cite{zhang2024gslrm, tang2024lgm, xu2024grm} effectively learn complex spatial relationships and 3D representations in a data-driven manner, in contrast to traditional geometry-based pipelines~\cite{schoenberger2016sfm,hartley2003multiple}. These approaches often reformulate 3D reconstruction as a multi-view 2D reasoning problem, where a set of multi-view input images is tokenized and processed as a sequence of tokens. This sequence-to-sequence formulation allows transformer architectures to model global geometric relationships across views and infer consistent scene representations. However, as each high-resolution image contributes a large number of tokens, the sequence length grows rapidly with the number of input views, leading to substantial computational and memory overhead due to the quadratic complexity of self-attention. This scalability issue poses a major challenge in applying transformer-based architectures to large-scale 3D reconstruction tasks involving many high-resolution images.

Motivated by these limitations, a growing body of work has introduced architectural designs to improve the efficiency and scalability of multi-view transformer models. Long-LRM~\cite{ziwen2025longlrm} proposed to integrate the transformer and bidirectional Mamba~\cite{mamba2} blocks. While improving efficiency through Mamba's linear-complexity design, its expressive capacity remains limited compared to the transformer's self-attention. iLRM~\cite{kang2025ilrm} employs a compact scene representation, which enables full attention across all input views. Although it achieves promising reconstruction quality and improved scalability, its reliance on global attention introduces a computational bottleneck as the number of input views increases. LVT~\cite{imtiaz2025lvt} adopts a local-view attention mechanism to improve scalability, where each input view attends only to nearby views. However, because attention is restricted to neighboring views, global 3D consistency is achieved only indirectly through multiple layers of local interactions. Moreover, defining neighborhood relationships between input views is itself a challenging problem, and the reliance on known camera poses further limits the flexibility and applicability of the architecture.

In this work, we introduce \textit{Multi-view Pyramid Transformer} (MVP), a scalable multi-view pyramid transformer architecture that substantially improves both scalability and overall performance compared to existing approaches. The key intuition behind the proposed architecture is \textit{`looking broader to see the whole, looking finer to see the details'}, which has been successfully adopted in many contemporary architectures.
Early convolutional networks~\cite{simonyan2014very, he2016deep, lin2017fpn} evolve feature representations from fine-grained spatial feature maps in the initial layers to coarse, semantically rich maps in the later layers, capturing global context while reducing computational cost. A comparable architectural design also emerges in modern transformer architectures, such as the Swin Transformer~\cite{liu2021Swin, liu2022videoswintransformer}, which progressively reduces the spatial and temporal resolutions of image tokens across layers. Building on this fine-to-coarse design philosophy, we introduce \textit{Dual Attention Hierarchy}, which enables scalable multi-view reasoning by structuring attention hierarchically along two complementary dimensions: inter-view and intra-view hierarchy.

\noindent (1) In the inter-view hierarchy, the attention window progressively expands to cover a broader range of input views, enabling the model to reason from neighboring inter-view relationships to global scene context. (2) In the intra-view hierarchy, the spatial image tokens are gradually merged across network layers to enlarge the receptive field within each input view, allowing the model to capture visual cues across multiple scales from fine details to coarse structures. In the early layers, the model operates on narrower view windows and fine-grained image tokens to efficiently extract local geometric details, while the later layers adopt wider view windows and coarse-grained image tokens to integrate broader contextual information. The two hierarchies operate in complementary directions, local-to-global in the inter-view hierarchy and fine-to-coarse in the intra-view hierarchy, achieving a balanced trade-off between computational efficiency and representational richness.

Beyond computational efficiency, there is another fundamental reason motivating our architectural design. Although full global attention theoretically enables all-to-all interactions across input views, it often fails to generalize in long-context settings~\cite{hyeon2023scratching, liu2024lost, ye2024differential}. As the number of input views increases, the attention distribution becomes diluted and unstable, leading to degraded correspondence learning and inconsistent feature alignment. In multi-view 3D reconstruction, this limitation manifests as diminishing performance gains when scaling to more input views, reflecting the difficulty of maintaining consistent geometric correspondence under long-range attention. In contrast, our \textit{Dual Attention Hierarchy} prevents the number of tokens involved in attention from growing excessively across layers, thereby avoiding attention dilution, stabilizing optimization in long-context regimes, and ensuring consistent multi-view reasoning as the number of input views increases.

We validate the effectiveness of the proposed architecture within the feed-forward 3D Gaussian Splatting (3DGS) framework. The proposed model efficiently processes up to 128 input views at a resolution of 960$\times$540 in under one second on a single H100 GPU, significantly outperforming the previous approaches in both inference speed and reconstruction quality (Fig.~\ref{fig:teaser}). Remarkably, it often achieves superior reconstruction quality to optimization-based methods such as 3D-GS in both sparse and dense-view settings, even without any additional post-processing. These results demonstrate that our method not only ensures high scalability but also delivers enhanced reconstruction performance compared to existing feed-forward 3DGS models. To summarize, our main contributions are as follows,
\begin{itemize}
\item We propose MVP, a novel scalable multi-view pyramid transformer architecture that significantly improves both efficiency and performance.

\item We introduce a dual hierarchical attention mechanism that jointly operates inter-view and intra-view attentions, enabling efficient multi-view reasoning.

\item We conduct extensive experiments demonstrating that our method achieves state-of-the-art performance while maintaining remarkable efficiency and scalability across diverse datasets and view configurations.

\item We analyze the proposed architecture from multiple perspectives, revealing how our design effectively achieves both global and local 3D consistency.

\end{itemize}
\vspace{-1mm}
\section{Related Work}
\vspace{-1mm}
\label{sec:related}

\subsection{Multi-view transformers}
\vspace{-1mm}
Recent progress in multi-view transformers, with DUSt3R~\cite{wang2024dust3r} serving as a notable milestone, has established a powerful paradigm for learning 3D geometry directly from multi-view imagery. This progress has spurred a range of methods that tackle different facets of 3D reconstruction. The first category comprises 3D geometry reconstruction approaches~\cite{wang2024dust3r, wang2025vggt, wang2025pi3, Yang_2025_Fast3R} that jointly estimate camera poses and scene structure in a feed-forward manner, often predicting dense point maps or depth fields in a single pass. The second category includes large view synthesis models~\cite{flynn2024quark, jin2025lvsm, jiang2025rayzer} that prioritize photorealistic novel view generation, typically emphasizing visual fidelity and scalability rather than explicitly recovering 3D geometry. The third category encompasses large reconstruction models~\cite{zhang2024gslrm, wei2024meshlrm, ziwen2025longlrm, kang2025ilrm, imtiaz2025lvt}, which leverage high-capacity neural networks and explicit or implicit 3D representations~\cite{mildenhall2020nerf, kerbl20233dgs} to reconstruct large-scale scenes while also enabling photorealistic novel view synthesis.
Although the aforementioned works differ in goals and architectures, they largely incorporate multi-view attention, a mechanism that naturally facilitates correspondence reasoning across input views, an essential component of 3D reconstruction. 
However, multi-view attention remains computationally expensive and struggles to scale to a large number of input views. 
In this work, we propose multi-view pyramid attention architecture that advances the field toward more scalable large-scale 3D reconstruction.

\subsection{Efficient sequence models}
\vspace{-1mm}
To model long-range dependencies, transformer-based architectures have become the dominant paradigm across diverse domains, including natural language processing~\cite{vaswani2017attention, brown2020language}, computer vision~\cite{dosovitskiy2020vit, liu2021Swin}, and video understanding~\cite{arnab2021vivit, bertasius2021timesformer}.
However, the quadratic complexity of global self-attention poses significant computational and memory challenges, particularly for high-resolution inputs or long sequences.
Efforts to improve efficiency generally follow two directions: (1) modifying the attention mechanism, by restricting attention to sparse patterns or local windows~\cite{child2019sparse, zaheer2020bigbird, beltagy2020longformer, liu2021Swin, zhang2023efficientnlp} or linearizing to sub-quadratic complexity~\cite{katharopoulos2020linear, gu2024mamba, mamba2}; and (2) optimizing the token representation, by compressing or merging tokens~\cite{bolya2022tome, yang2025visionzip, xia2025tokenskip, ziwen2025longlrm} to reduce sequence length without losing essential information.
Previous work~\cite{zhang2020featurepyramid, hatamizadeh2021swinunet, hu2025ultragen} adopts hierarchical architectures to enhance efficiency, but their hierarchies are confined to individual spatial domains. Moreover, temporal downsampling~\cite{liu2022videoswintransformer} is not straightforward in the multi-view setting, so we instead reduce the spatial token resolution to accommodate more views.
Meanwhile, Long-LRM~\cite{ziwen2025longlrm} and iLRM~\cite{kang2025ilrm} introduce efficient multi-view transformers for 3D reconstruction, but they still rely on the same full global attention mechanism across all input views, leading to high computational cost and limited scalability.

The proposed approach explores a different design philosophy by introducing a \textit{Dual Attention Hierarchy} that operates jointly along view and spatial dimensions. This hierarchical formulation enables the model to capture both local geometric cues and global multi-view relationships in a scalable and computationally efficient manner. 
Different from~\cite{liu2022videoswintransformer}, which operates in a temporal video setting, our model works in a multi view setting and progressively increases the number of views, analogous to temporal length in videos, whereas they do not progressively extend the temporal dimension in different stages. 
We also employ Alternative Attention~\cite{wang2025vggt} to further improve efficiency when processing more than two views simultaneously.

\vspace{-1mm}
\section{Method}
\vspace{-1mm}
\label{sec:method}

\begin{figure*}[!h]
    \centering
    \includegraphics[width=1.0\textwidth]{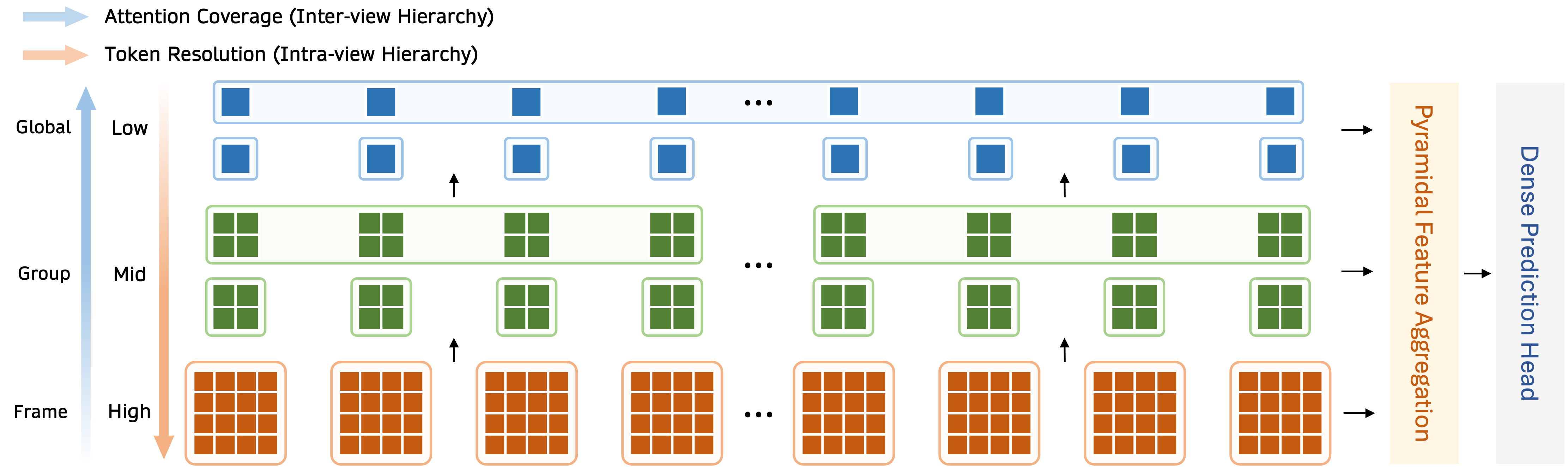}    
    \vspace{-2mm}    
    \caption{Architecture Overview. Given tokenized inputs, our model applies a three stage hierarchy of alternating attention blocks, varying in both self-attention coverage and token resolution. A Pyramidal Feature Aggregation module fuses the outputs from all stages, which are then passed to a final head for dense prediction.} 
    \label{fig:overview}
    \vspace{-4mm}
\end{figure*}

We introduce MVP, a scalable multi-view dense prediction transformer architecture capable of processing more than one hundred input-view images in under one second. The core of our model is the \textit{Dual Attention Hierarchy}, which progressively integrates contextual and spatial information through inter-view and intra-view hierarchies. While our formulation is general and applicable to a wide range of multi-view tasks, in this work, we instantiate it within the feed-forward 3D Gaussian Splatting (3DGS) framework, enabling efficient and high-quality 3D reconstruction. 

\vspace{-1mm}
\subsection{Overall Architecture}
\vspace{-1mm}
\label{subsec:overall_arch}
Our MVP transformer architecture (\cref{fig:overview}) comprises an input tokenizer (\cref{subsec:input_encoding}) followed by a three-stage hierarchy of attention blocks (\cref{subsec:inter_hierarchy}), with token reduction modules interleaved between each stage (\cref{subsec:intra_hierarchy}). A pyramidal feature aggregation module then collects the outputs from each stage, fusing them into a unified feature (Sec.\ref{subsec:feat_agg}). Finally, a decoder processes the fused representation to produce the dense multi-view predictions (\cref{subsec:output_decoding}).

\vspace{-1mm}
\subsection{Input Encoding}
\vspace{-1mm}
\label{subsec:input_encoding}
Given a set of $N$ input images $\{I_i\}_{i=1}^{N}$, where $I_i \in \mathbb{R}^{H\times W \times3}$, we begin by tokenizing the input views, which are subsequently processed by a series of transformer layers. To incorporate camera geometry, we encode the known camera poses as 9D Plücker ray map, $P_i \in \mathbb{R}^{H\times W \times 9}$, where each ray is represented by concatenating its origin, direction, and their cross product~\cite{zhang2025test}. Then, each input image $I_i$ is concatenated with its corresponding ray map $P_i$ along the channel dimension, forming a 12-channel posed-image tensor $\tilde{I}_i \in \mathbb{R}^{H \times W \times 12}$. This posed image tensor is then patchified into non-overlapping patches of size $p$ through a linear projection layer. For each input view, we further append four register tokens~\cite{darcet2023vision, wang2025vggt, simeoni2025dinov3}, resulting in a total of $N(HW/p^2+4)$ tokens that serve as input to the subsequent transformer stages.
\vspace{-1mm}
\subsection{Dual Attention Hierarchy}
\vspace{-1mm}
\label{subsec:attn_hierarchy}

MVP employs a dual-level attention hierarchy consisting of inter-view and intra-view attention mechanisms. In our current implementation, this hierarchy comprises three stages, progressively integrating multi-view contextual information from local to global and visual cues from fine- to coarse-grained scales in a computationally efficient manner.

\vspace{-1mm}
\subsubsection{Inter-view Attention Hierarchy}
\label{subsec:inter_hierarchy}

We introduce a group-wise self-attention mechanism that attends to tokens within predefined groups of views, serving as an efficient intermediate stage between purely local (frame-wise) and fully global attention. This formulation enables scalable multi-view interaction while preserving global consistency. Our framework also generalizes the Alternating-Attention (AA) module~\cite{wang2025vggt} as a special case, encompassing frame-wise, group-wise, and global self-attention within a unified framework.

\noindent \textbf{Stage 1}: The first stage focuses on extracting spatial features and local geometric details from each frame independently, employing only frame-wise attention.

\noindent \textbf{Stage 2}: The second stage begins to reason about inter-view relationships utilizing group-wise self-attention. First, we apply a grouping operator, $\text{group}(\cdot): \mathbb{R}^{Nhw \times d} \rightarrow \mathbb{R}^{\frac{N}{M} \times Mhw \times d}$, where $M$ is the number of views within a group ($h, w$ is the downsampled token resolution, see \cref{subsec:intra_hierarchy}). In this work, the grouping operator simply partitions the $N$ views into $\frac{N}{M}$ consecutive groups by the locality of frame indices. Then, we perform frame-wise self-attention, followed by a group-wise self-attention.
\vspace{-1mm}
\begin{equation}
\begin{aligned} 
&G \leftarrow \text{group}(T), \\
&{G}_{i,j} \leftarrow \text{self-att}(G_{i,j}) \quad\hspace{-1mm} \forall i,j \quad\hspace{-1mm} \text{(frame-wise attention)},\\ 
&{T}_i \leftarrow \text{self-att}({G}_i) \quad\hspace{-1mm} \forall i \quad\hspace{-1mm} \text{(group-wise attention)},
\end{aligned} 
\vspace{-1mm}
\end{equation}
where $T \in \mathbb{R}^{Nhw \times d}$ represents all input tokens, $G_{i,j} \in \mathbb{R}^{hw \times d}$ denotes the tokens that correspond to the $j$-th image in the $i$-th group, and $G_{i} \in \mathbb{R}^{Mhw \times d}$ is every token in the $i$-th group (we omit the register tokens for brevity).

\noindent \textbf{Stage 3}: At the final stage, the model integrates information across all views to form a global understanding of the scene and construct a fully coherent 3D representation. In this stage, the architecture simplifies to the AA module, where all views belong to the same single group ($M==N$).

\subsubsection{Intra-view Attention Hierarchy}
\label{subsec:intra_hierarchy}
The intra-view fine-to-coarse hierarchy operates along the spatial dimensions of each input frame. This scaling strategy enables the model to transition from fine-grained local reasoning in early layers to coarse-grained global reasoning in later layers within each image. We implement this using a single convolution layer between stages that performs spatial downsampling and channel up-projection simultaneously. At each stage, the number of image tokens is reduced by a factor of four ($h\rightarrow \frac{h}{2}$ and $w \rightarrow \frac{w}{2}$), enlarging the effective receptive field of each token. Meanwhile, the token embedding dimension is doubled, increasing feature capacity to encode information from this expanded spatial region. This progressive token reduction not only forms a feature pyramid, but also allows larger group-wise attention as the spatial resolution decreases.

\vspace{-1mm}
\subsection{Pyramidal Feature Aggregation}
\vspace{-1mm}
\label{subsec:feat_agg}

To perform dense multi-view prediction, we introduce a Pyramidal Feature Aggregation (PFA) module that integrates multi-scale and multi-stage features produced by our dual attention hierarchy. While conceptually related to the Dense Prediction Transformer~\cite{ranftl2021vision}, the proposed PFA is specifically tailored to the hierarchical structure of our architecture, enabling effective fusion across the three stages.

Concretely, multi-view feature tokens from all stages are collected and fused through a top-down refinement process. First, we reshape the tokens from all stages into spatial feature maps, then a convolutional layer is applied to project them into a shared high-dimensional space while preserving spatial resolution. The resulting feature maps are progressively upsampled and fused with those from preceding stages through residual convolutional fusion blocks, allowing the model to combine coarse global context with fine local details. Finally, the aggregated feature tensor is rearranged back into a token sequence and passed to the decoder for dense 3D Gaussian prediction. More formally, the aggregated feature $F$ can be obtained as follows,
\vspace{-1.2mm}
\begin{equation}
\begin{aligned} 
F = \text{fuse}(\text{up}(\text{fuse}(\text{up}(F^{(3)}) + F^{(2)})) + F^{(1)}),
\end{aligned} 
\vspace{-1.2mm}
\end{equation}
where $F^{(1)}, F^{(2)}, F^{(3)}$ are the reshaped output feature maps from each stage, $\text{up}(\cdot)$ and $\text{fuse}(\cdot)$ denote an upsampling layer and feature fusion, respectively.

\begin{figure*}[!h]
    \centering
    \includegraphics[width=1.0\textwidth]{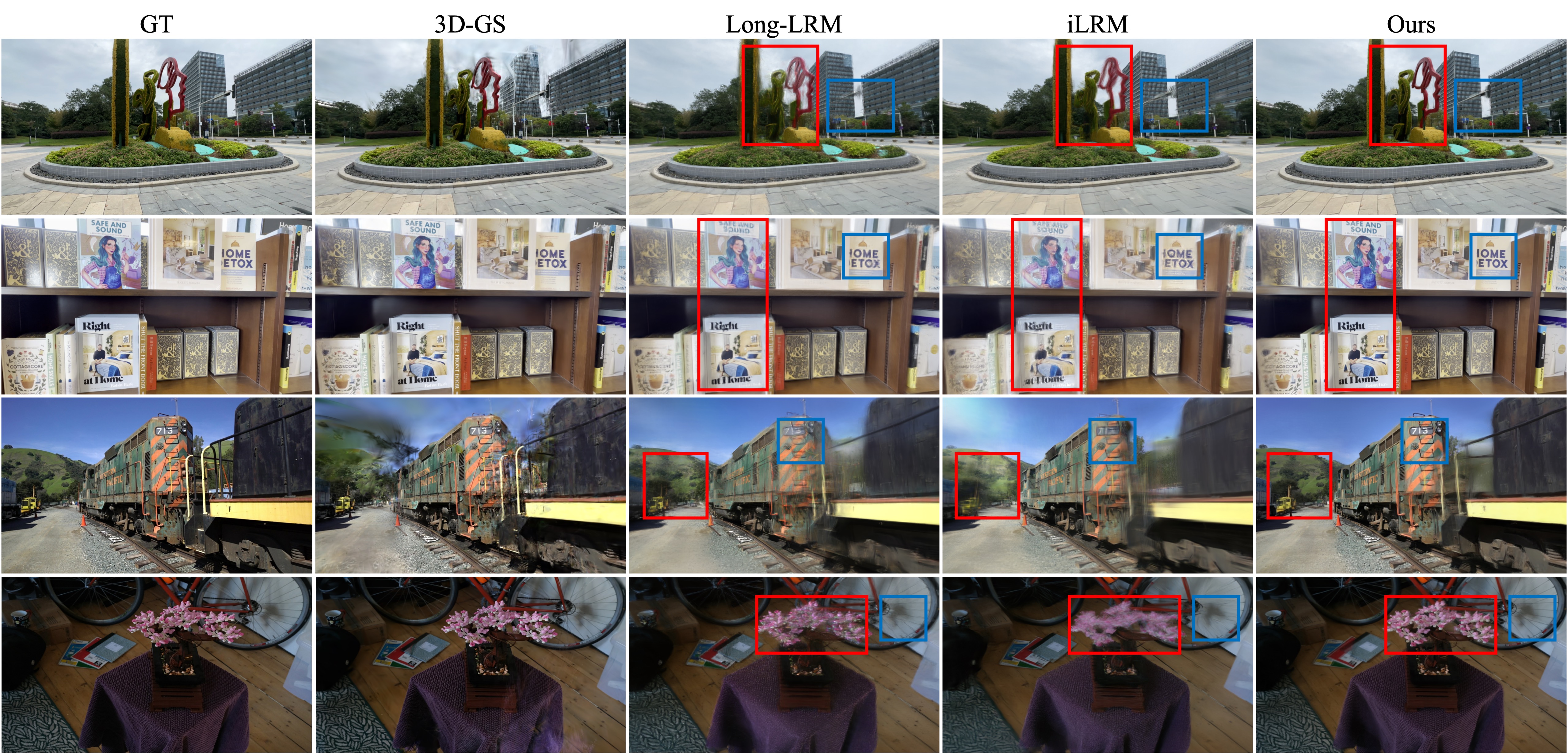} 
    \vspace{-6mm}    
    \caption{Qualitative results on the DL3DV (top two rows), Tanks\&Temples (third row), and Mip-NeRF360 (bottom row). For a fair and reliable comparison, we evaluate all methods with 32 input views, matching the training setup used for other feed-forward baselines.}  
    \vspace{-1mm}
    \label{fig:qual_main}
\end{figure*}

\begin{table*}[!h]
    \centering
    \resizebox{\linewidth}{!}{
        \setlength{\tabcolsep}{2pt}
        \renewcommand{\arraystretch}{1.0}
        \begin{tabular}{@{}lcccccccccccccccc@{}}
        \toprule
        \multirow{2}{*}{Method} & \multicolumn{4}{c}{16 views} & \multicolumn{4}{c}{32 views} & \multicolumn{4}{c}{64 views} & \multicolumn{4}{c}{128 views} \\
        \cmidrule(lr){2-5}\cmidrule(lr){6-9}\cmidrule(lr){10-13}\cmidrule(lr){14-17}          & PSNR $\uparrow$ & SSIM $\uparrow$ & LPIPS $\downarrow$ & Time (s) $\downarrow$ & PSNR $\uparrow$ & SSIM $\uparrow$ & LPIPS $\downarrow$ & Time (s) $\downarrow$ & PSNR $\uparrow$ & SSIM $\uparrow$ & LPIPS $\downarrow$ & Time (s) $\downarrow$ & PSNR $\uparrow$ & SSIM $\uparrow$ & LPIPS $\downarrow$ & Time (s) $\downarrow$ \\ 
         \midrule
        3D-GS$_{30k}$~\cite{kerbl20233dgs} & 21.48 & 0.753 & 0.252 & 8min & 24.43 & 0.827 & 0.191 & 8min & 27.34 & 0.883 & 0.146 & 8min & 29.43 & 0.914 & 0.123 & 8min \\
        \midrule
        Long-LRM~\cite{ziwen2025longlrm} & 21.05 (22.66) & 0.708 (0.740) & 0.297 (0.292) & 0.50 & 23.97 & 0.778 & 0.267 & 0.84 & 23.60 & 0.789 & 0.260 & 2.08 & 21.24 & 0.739 & 0.308 & 6.39 \\
        iLRM~\cite{kang2025ilrm} & 21.92 (22.91) & 0.748 (0.766) & 0.316 (0.295) & 0.19 & 24.30 & 0.803 & 0.256 & 0.53 & 24.44 & 0.819 & 0.240 & 1.66 & 22.98 & 0.807 & 0.249 & 5.61 \\
        Ours & \best{23.76} & \best{0.798} & \best{0.239} & \best{0.09} & \best{25.96} & \best{0.847} & \best{0.187} & \best{0.17} & \best{27.73} & \best{0.881} & \best{0.154} & \best{0.36} & \best{29.02} & \best{0.903} & \best{0.134} & \best{0.77} \\
        \bottomrule
        \end{tabular}
    }
    \vspace{-3mm}    
    \caption{
    Quantitative comparisons on the DL3DV dataset with varying numbers of input views. The values in parentheses $(\cdot)$, for the 16-view setting are taken from the original papers, where the models are trained with a fixed 16-view input. For all metrics, we report the results by re-evaluating the models from their 32-view checkpoints.}
    \label{tab:quantitative result on dl3dv}  
    \vspace{-3mm}        
\end{table*}    

\begin{table}[!h]
    \centering
    \resizebox{1.0\columnwidth}{!}{
        \setlength{\tabcolsep}{2pt}
        \renewcommand{\arraystretch}{1.1}
        \begin{tabular}{@{}lcccccccc@{}}
        \toprule        
        \multirow{2}{*}{Method} & \multicolumn{4}{c}{192 views} & \multicolumn{4}{c}{256 views} \\
        \cmidrule(lr){2-5}\cmidrule(lr){6-9} & PSNR $\uparrow$ & SSIM $\uparrow$ & LPIPS $\downarrow$ & Time (s) $\downarrow$ & PSNR $\uparrow$ & SSIM $\uparrow$ & LPIPS $\downarrow$ & Time (s) $\downarrow$ \\
        \midrule
        3D-GS$_{30k}$~\cite{kerbl20233dgs} & 30.14 & 0.924 & 0.116 & 8min & 30.39 & 0.926 & 0.114 & 8min \\
        \midrule
        Long-LRM~\cite{ziwen2025longlrm} & \multicolumn{8}{c}{OOM (GPU memory limit exceeded, 80\,GB)} \\
        iLRM~\cite{kang2025ilrm} & 21.68 & 0.788 & 0.264 & 11.91 & 20.63 & 0.767 & 0.281 & 20.92 \\
        Ours & \best{29.54} & \best{0.912} & \best{0.127} & \best{1.23} & \best{29.67} & \best{0.915} & \best{0.125} & \best{1.84} \\
        \bottomrule        
        \end{tabular}
    }
    \vspace{-1mm}    
    \caption{Quantitative comparisons on the DL3DV dataset with varying numbers of input views.}
    \label{tab:quantitative result on dl3dv2}      
    \vspace{-0mm}     
\end{table}

\begin{table}[!h]
    \centering
    \resizebox{1.0\columnwidth}{!}{
        \renewcommand{\arraystretch}{1.0}
        \begin{tabular}{lccccccc}
        \toprule
        \multirow{2}{*}{Method} & \multirow{2}{*}{Views} & \multicolumn{3}{c}{Tanks \& Temples} & \multicolumn{3}{c}{Mip-NeRF360} \\
        \cmidrule(lr){3-5}\cmidrule(lr){6-8}
        & & PSNR $\uparrow$ & SSIM $\uparrow$ & LPIPS $\downarrow$ & PSNR $\uparrow$ & SSIM $\uparrow$ & LPIPS $\downarrow$ \\
        \midrule       
        Long-LRM~\cite{ziwen2025longlrm} & \multirow{3}{*}{32} & 18.59 & 0.614 & 0.366 & 21.08 & 0.484 & 0.445 \\
        iLRM~\cite{kang2025ilrm} & & 18.58 & 0.631 & 0.385 & 21.09 & 0.495 & 0.466 \\
        Ours & & \best{19.54} & \best{0.708} & \best{0.277} & \best{22.21} & \best{0.587} & \best{0.355} \\
        \midrule       
        Long-LRM & \multirow{3}{*}{64} & 19.44 & 0.651 & 0.334 & 21.30 & 0.499 & 0.431 \\
        iLRM & & 19.82 & 0.692 & 0.318 & 21.60 & 0.522 & 0.444 \\
        Ours & & \best{21.24} & \best{0.761} & \best{0.221} & \best{23.72} & \best{0.656} & \best{0.302} \\
        \midrule

        Long-LRM & \multirow{3}{*}{128} & 18.47 & 0.613 & 0.375 & 19.82 & 0.484 & 0.457 \\
        iLRM & & 19.22 & 0.696 & 0.319 & 21.32 & 0.551 & 0.424 \\
        Ours & & \best{22.36} & \best{0.804} & \best{0.184} & \best{25.12} & \best{0.736} & \best{0.248} \\        
        \bottomrule
        \end{tabular}
    }
    \vspace{-1mm}    
    \caption{Quantitative comparisons on the Tanks\&Temples and Mip-NeRF360 under a zero-shot setting. For the 128 view setting on Mip-NeRF360, the \texttt{stump} and \texttt{treehill} scenes are excluded due to their limited number of frames.}
    \vspace{-5mm}    
    \label{tab:quantitative result on cross}    
\end{table}

\subsection{Output Decoding}
\vspace{-1mm}
\label{subsec:output_decoding}
Given the aggregated tokens, we apply a single linear layer for dense prediction. In our feed-forward 3DGS setting, each output pixel parameterizes a 3D Gaussian primitive with attributes $(\mu_j, s_j, q_j, \alpha_j, c_j)$ for position, scale, rotation (quaternion), opacity, and color, respectively. Following prior works~\cite{nowak2025vod, imtiaz2025lvt}, our model predicts spherical harmonic coefficients for both view-dependent color and opacity, enabling accurate modeling of appearance variations across viewing directions.

\subsection{Training Objective}
\vspace{-1mm}
\label{subsec:train_obj}
By rendering novel view images from the predicted Gaussian primitives using the target camera poses, we compare the rendered target view images $\hat{I}_i$ with ground-truth target view images $I_i$ to compute and minimize the loss. We use a weighted sum of MSE loss and perceptual loss.
\vspace{-1.2mm}
\begin{align}
  \mathcal{L}_\text{img} = \frac{1}{|\mathcal{T}|}\sum_{i \in \mathcal{T}}(\mathcal{L}_\text{MSE}(\hat{I}_{i}, I_{i}) + \lambda \mathcal{L}_\text{percept}(\hat{I}_{i}, I_{i})),
  \vspace{-1.2mm}
\end{align}
where $\mathcal{T}$ is a set of target view indices, and we set $ \lambda=0.2$ as the weighting factor throughout the paper. We also regularize view-dependent splat opacity to mitigate undesirable opacity artifacts across different viewpoints~\cite{imtiaz2025lvt}. Let $N_\mathcal{G}$ ($N_\mathcal{G}= NHW$ for per-pixel Gaussians) be the number of predicted Gaussian primitives, and we compute the regularizing factor 
$\mathcal{R}_\alpha$ as
\vspace{-1.3mm}
\begin{align}
\mathcal{R}_\alpha = \frac{1}{N_\mathcal{G}}\sum_{j=1}^{N_\mathcal{G}}|\sigma(\alpha_j \cdot \omega_j)|,
\vspace{-4mm}
\end{align}
where $\sigma(\cdot)$ is the Sigmoid function and $\omega_j$ the spherical harmonic basis from the randomly sampled per-pixel view direction. Our overall loss function is $\mathcal{L} = \mathcal{L}_\text{img} + \gamma\mathcal{R}_\alpha$, where we set $\gamma=0.001$ as the weighting factor.

\section{Experiments}
\vspace{-2mm}
\label{sec:experiments}

\textbf{Dataset.} We conduct experiments on the DL3DV dataset~\cite{ling2024dl3dv}, which consists of large-scale multi-view scenes spanning diverse indoor and outdoor environments. Each scene is processed using COLMAP~\cite{schoenberger2016sfm, schoenberger2016mvs} to obtain accurate camera poses, containing approximately 250–350 frames per scene. Following prior works~\cite{ziwen2025longlrm, kang2025ilrm}, we adopt the official training and benchmark splits provided by the dataset.  
We also evaluate the zero-shot generalization performance on the Tanks\&Temples ~\cite{Knapitsch2017tnt} and Mip-NeRF360~\cite{barron2022mipnerf360} datasets, which contain high-quality real-world scenes captured under challenging lighting and geometry conditions. For all datasets, we use images at a resolution of 960$\times$540 and evaluate with different numbers of input views, ranging from 16 to 256. We further evaluate under a low resolution setting (256$\times$256) on the RealEstate10K (RE10K) dataset~\cite{zhou2018re10k}, following the same train-test split and indices protocol in~\cite{charatan2024pixelsplat, kang2025ilrm}.

\noindent\textbf{Implementation details.} We utilize 2 frame-wise, 4 group-wise, and 8 global attention blocks for each stage. Tokens initially correspond to 8$\times$8 patches and are progressively merged to 16$\times$16 and 32$\times$32 across stages, with embedding sizes 256, 512, and 1024. We also apply PRoPE~\cite{li2025prope} to encode cameras as relative positional signals, providing geometry aware conditioning for our attention layers.

\noindent\textbf{Training details.} We adopt a three stage training schedule, starting at low resolution with a fixed number of views and progressing to high resolution with mixed input view counts. In the first stage, we train the model at 480$\times$256 with 32 input views and 12 target views. In the second stage, we switch to 960$\times$540 while keeping 32 input views and reducing to 6 target views. In the final stage, we continue to train at 960$\times$540 with varying numbers of input views, selecting the number of target views according to GPU memory constraints. In this stage, we freeze the parameters of the first two architectural stages (frame- and group-wise blocks) and update only the global modules. Each training stage takes approximately 4, 3, and 2 days, respectively, using 32 H100 GPUs. 

\subsection{Results}
\vspace{-1mm}
\label{sec:result}
Throughout the manuscript, inference times are measured on a single NVIDIA H100 GPU with FlashAttention3~\cite{shah2024flashattention}.

\noindent\textbf{Comparison.} Tab.~\ref{tab:quantitative result on dl3dv} and~\ref{tab:quantitative result on dl3dv2} report quantitative comparisons on the DL3DV benchmark dataset. We compare our method with the optimization based 3D Gaussian Splatting~\cite{kerbl20233dgs} with 30K optimization steps and two recent feed-forward reconstruction methods, Long-LRM~\cite{ziwen2025longlrm} and iLRM~\cite{kang2025ilrm}. Across all evaluated settings from 16 to 256 input views, our method consistently outperforms previous single-pass models by a large margin in both reconstruction quality and inference efficiency. Note that, unlike our training scheme, the baselines are trained only with 32 input views, and do not exhibit scalability as the number of views increases. Also, in the very dense 256-view setting, our approach remains within 0.7 dB PSNR of the optimization-based baseline, but over 250$\times$ faster, while Long-LRM fails to run due to out-of-memory (OOM) issues. In Tab.~\ref{tab:quantitative result on cross}, we report generalization performance with Tanks\&Temples and Mip-NeRF360 dataset. 
For all tested view counts (32, 64, and 128), our method consistently exceeds the baselines on all the evaluation metrics and the gap increases as the number of input views increases.

Tab.~\ref{tab:quantitative result on re10k} summarizes the results on the low resolution ($256 \times 256$) RE10K dataset with 4 and 8 input views. We adopt prior methods~\cite{wang2025clift, kang2025ilrm} as baselines. For \cite{wang2025clift}, we select the model that performs best among the 4-view configurations.
Our coarse variant of patch sizes (8, 16, 32) already outperforms the baselines, while our fine variant of patch sizes (4, 8, 16) further improves the rendering quality. Note that our models are trained with varying numbers of input views, unlike the baselines.

\begin{table}[!h]
    \centering
    \resizebox{1.0\columnwidth}{!}{
        \begin{tabular}{lcccccc}
        \toprule
        \multirow{2}{*}{Method} & \multicolumn{3}{c}{4 views} & \multicolumn{3}{c}{8 views} \\
        \cmidrule(lr){2-4}\cmidrule(lr){5-7}   
         & PSNR $\uparrow$ & SSIM $\uparrow$ & LPIPS $\downarrow$ & PSNR $\uparrow$ & SSIM $\uparrow$ & LPIPS $\downarrow$ \\
        \midrule
        CLiFT~\cite{wang2025clift}  & 30.13 & 0.916 & 0.095 & 29.68 & 0.910 & 0.092 \\
        iLRM~\cite{kang2025ilrm}  & 30.37 & 0.923 & 0.095 & 28.90 (31.57) & 0.920 (0.935) & 0.102 (0.082) \\
        Ours-coarse & 30.56 & 0.925 & 0.087 & 31.78 & 0.939 & 0.077 \\
        Ours-fine & \best{32.12} & \best{0.941} & \best{0.077} & \best{33.40} & \best{0.952} & \best{0.066} \\

        \bottomrule
        \end{tabular}
    }
    \vspace{-3mm}        
    \caption{Quantitative comparisons on the RE10K dataset. Baselines are evaluated using 4-view models, and values in parentheses $(\cdot)$ are from the models trained for that specific view setting.}
    \vspace{-3mm}    
    \label{tab:quantitative result on re10k}    
\end{table}

\begin{figure*}[!h]
    \centering
    \includegraphics[width=1.0\textwidth]{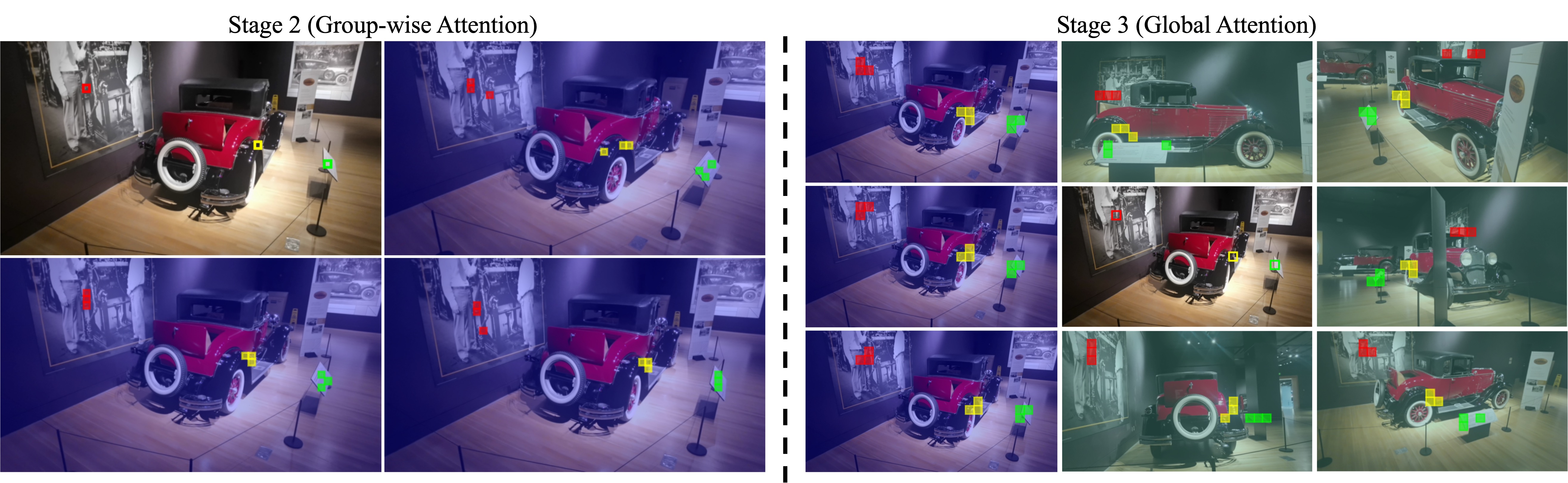}    
    \vspace{-4.9mm}
    \caption{Attention visualization. For colored query patches (\textcolor[HTML]{C02A1E}{red}, \textcolor[HTML]{D1D147}{yellow}, \textcolor[HTML]{62CF42}{green}) in the reference view, we highlight top-3 attended tokens: on the left, tokens attended within the group (blue overlay), and on the right, tokens attended within and outside the group (green overlay).}  
    \label{fig:att}
\end{figure*}

\noindent\textbf{Attention visualization.} To investigate how our method achieves both local (group) and global multi-view consistency, we visualize the learned attention patterns across views and stages, highlighting how the model adaptively attends to geometrically consistent regions within group and across the entire views. Using the first frame as the reference view, we first select three query patches. For each query patch, we then visualize its top-3 attended tokens from other views in different stages. At stage 2, we show tokens within the same group, while at stage 3, we show tokens from both within and outside the group. As shown in Fig.~\ref{fig:att}, the group-wise attention over fine tokens effectively captures local spatial correspondence, while the global attention over coarse tokens still focuses on semantically and geometrically consistent regions across distant views.

\noindent\textbf{Longer context generalization.} Tab.~\ref{tab:longer} evaluates how well each model generalizes to longer input contexts. All baselines, including our method, are only trained with 32 input views and then tested on 32 (seen), 40, and 48 views. Our approach not only achieves the best evaluation metrics at the training length, but also continues to improve most when moving from 32 to 40 and 48 views, whereas Long-LRM and iLRM show much smaller gains and tend to saturate. At the same time, the inference time of our model increases only slightly with more views and remains substantially lower than the baselines. These results support our claim that the proposed hierarchical design alleviates attention dilution in long context regimes and maintains stable multi-view reasoning when extrapolating to longer input contexts beyond the training range.

\begin{table*}[!h]
    \centering
    \resizebox{\linewidth}{!}{
        \renewcommand{\arraystretch}{0.90}    
        \begin{tabular}{lcccccccccccc}
        \toprule        
        \multirow{2}{*}{Method} & \multicolumn{4}{c}{32 views (Trained)} & \multicolumn{4}{c}{40 views (Unseen)} & \multicolumn{4}{c}{48 views (Unseen)} \\
        \cmidrule(lr){2-5}\cmidrule(lr){6-9}\cmidrule(lr){10-13}
        & PSNR $\uparrow$ & SSIM $\uparrow$ & LPIPS $\downarrow$ & Time (s) $\downarrow$ & PSNR $\uparrow$ & SSIM $\uparrow$ & LPIPS $\downarrow$ & Time (s) $\downarrow$ & PSNR $\uparrow$ & SSIM $\uparrow$ & LPIPS $\downarrow$ & Time (s) $\downarrow$ \\
        \midrule
        Long-LRM~\cite{ziwen2025longlrm}  & 23.97 & 0.778 & 0.267 & 0.84 & 24.18 (0.21) & 0.787 & 0.260 & 1.05 (0.21) & 24.30 (0.33) & 0.797 & 0.252 & 1.38 (0.54) \\
        iLRM~\cite{kang2025ilrm} & 24.30 & 0.803 & 0.257 & 0.53 & 24.54 (0.24) & 0.811 & 0.248 & 0.76 (0.23) & 24.78 (0.48) & 0.820 & 0.240 & 1.04 (0.51) \\
        Ours & \best{25.88} & \best{0.845} & \best{0.188} & \best{0.17} & \best{26.36} (\textcolor{darkgreen}{0.48}) & \best{0.855} & \best{0.178} & \best{0.22} (\textcolor{darkgreen}{0.05}) & \best{27.06} (\textcolor{darkgreen}{1.18}) & \best{0.870} & \best{0.164} & \best{0.26} (\textcolor{darkgreen}{0.09}) \\
        \bottomrule
        \end{tabular}        
        }
    \vspace{-2.5mm}
    \caption{Longer context generalization.  values in parentheses $(\cdot)$ denote the improvements over 32 views.
    }
    \label{tab:longer}
\end{table*}

\begin{table*}[!h]
    \centering
    \renewcommand{\arraystretch}{1}
    \resizebox{\linewidth}{!}{
        \renewcommand{\arraystretch}{0.98}    
        \begin{tabular}{lccccccccc}
        \toprule
        Method & Stage 1 & Stage 2 & Stage 3 & Patch Size & Hidden Dim. & Feature Agg. & PSNR $\uparrow$ & SSIM $\uparrow$ & LPIPS $\downarrow$\\
        \midrule
        Baseline & Frame(2) & Group(4) & Global(8) & (8, 16, 32) & (128, 256, 512) & o & 22.79 & 0.733 & 0.235\\
        \midrule
        w/o Feature Agg. & Frame(2) & Group(4) & Global(8) & (8, 16, 32) & (128, 256, 512) & x & 21.58 & 0.646 & 0.340\\
        \midrule
        \multirow{2}{*}{w/o Group-attn.}
        & Frame(2) & Frame(4) & Global(8) & \multirow{2}{*}{(8, 16, 32)} & \multirow{2}{*}{(128, 256, 512)} & \multirow{2}{*}{o} & 22.53 & 0.720 & 0.247 \\
        & Frame(2) & Global(4) & Global(8) & & & & 22.94 & 0.739 & 0.235 \\
        \midrule
        w/o Inter-view Hierarchy &  Global(2) & Global(4) & Global(8) & (8, 16, 32) & (128, 256, 512) & o & 22.94 & 0.739 & 0.236 \\
        w/o Intra-view Hierarchy & Frame(2) & Group(4) & Global(8) & (8, 8, 8) & (512, 512, 512) & o & 22.83 & 0.732 & 0.249\\
        w/o Dual Hierarchy (p=8) & Global(2) & Global(4) & Global(8) & (8, 8, 8) & (512, 512, 512) & o & 23.20 & 0.747 & 0.241\\
        w/o Dual Hierarchy (p=16) & Global(2) & Global(4) & Global(8) & (16, 16, 16) & (512, 512, 512) & o & 21.80 & 0.651 & 0.341\\        
        \midrule
        Reversed Hierarchy & Global(8) & Group(4) & Frame(2) & (32, 16, 8) & (512, 256, 128) & o & 18.95 & 0.442 & 0.555 \\
        \bottomrule
        \end{tabular}        
    }
    \vspace{-1.7mm}
    \caption{Ablations. We train all variants for 100K iterations at a reduced resolution (256$\times$256) on the DL3DV dataset.}
    \label{tab:ablations}
    \vspace{-3mm}   
\end{table*}

\begin{figure}[!h]
    \centering
    \begin{subfigure}[!h]{0.48\columnwidth}
        \centering
        \includegraphics[width=\linewidth]{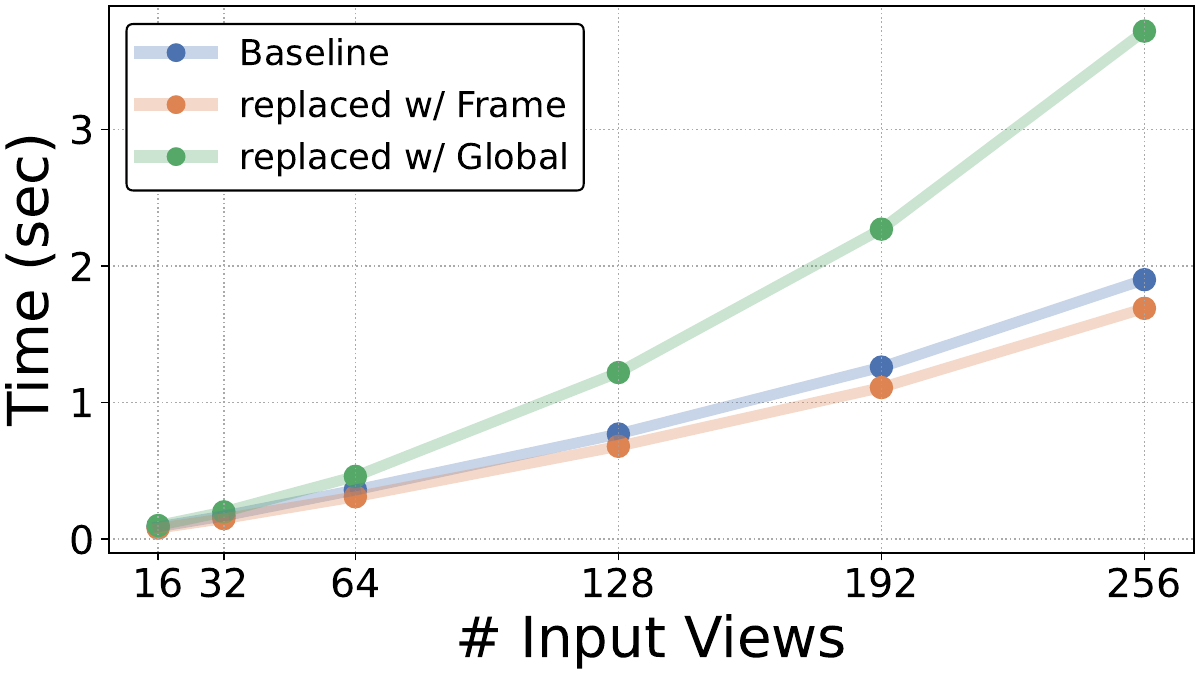}
        \caption{}
        \label{fig:time_wo_group}
    \end{subfigure}
    \begin{subfigure}[!h]{0.48\columnwidth}
        \centering
        \includegraphics[width=\linewidth]{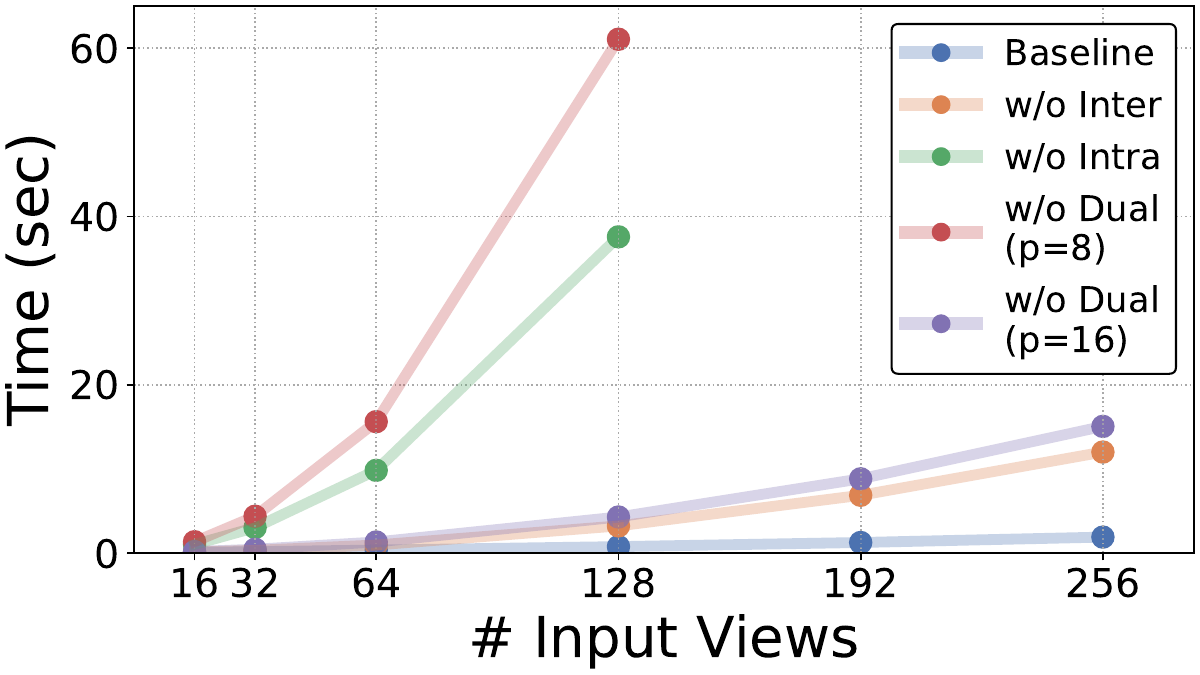}
        \caption{}
        \label{fig:time_wo_dual}
    \end{subfigure}
    \vspace{-2mm}    
    \caption{Ablation and inference time analysis of MVP's components. (a) Inference time analysis of group-attention mechanism. (b) Inference time analysis of the dual hierarchy components.}
    \vspace{-0mm}
\end{figure}

\section{Ablation and Analysis}
\vspace{-1mm}
\label{sec:ablation}

We conduct ablation experiments in Tab.~\ref{tab:ablations} to validate the proposed architectural components, with a particular focus on reducing computational cost and improving scalability while maintaining comparable reconstruction quality. 

\noindent\textbf{1) Feature aggregation.} To validate our Pyramidal Feature Aggregation (PFA), we tested a variant that decodes only from the final, coarse feature map. This variant showed degraded reconstruction quality, highlighting the importance of the PFA's multi-scale fusion for generating fine-grained representations, especially comparing LPIPS metric.

\noindent\textbf{2) Group-wise attention.} We ablate the group-wise attention by constructing two variants: one where all group-wise blocks are replaced with frame-wise blocks, and another where they are replaced with global blocks. The results exhibit a clear computation–performance trade-off: while global attention attains the best accuracy at the cost of higher complexity, our group-wise attention recovers most of the performance gains with lower computational overhead. Moreover, as our group size is fixed to four views, the computational cost of the global variant grows much more rapidly as the number of input views increases (Fig.~\ref{fig:time_wo_group}).

\noindent\textbf{3) Dual hierarchy.} We further analyze our dual-attention hierarchy by comparing variants without the intra-view hierarchy, without the inter-view hierarchy, and without both components. Note that our variant without both components is analogous to a modified version of VGGT~\cite{wang2025vggt}, using a linear patch-embedding layer instead of a DINO~\cite{oquab2023dinov2} encoder and omitting camera tokens. As shown in Fig.~\ref{fig:time_wo_dual}, the hierarchies are critical for scalability. In particular, for 256 input images, removing the inter-view hierarchy results in more than 6$\times$ higher latency. Moreover, the variants without the intra-view hierarchy and without both intra- and inter-view hierarchies encounter out-of-memory errors at 256 views; even at 64 input images, they are approximately 50$\times$ and 80$\times$ slower than our method, respectively. Notably, compared to the variant that uses a patch size of 16 without dual hierarchy, our method achieves a lower computational cost while permitting a finer, higher-performance patch size compared to non-hierarchical alternatives.

\noindent\textbf{4) Reversed hierarchy.} For the reversed variant, we change the intra-view hierarchy to apply attention in the order of global, group, and frame, and modify the inter-view branch to start from the lowest-resolution features and progressively upsample. This design yields clearly lower performance, confirming the effectiveness of our design.

\noindent\textbf{5) Redundancy pruning.} We prune Gaussian primitives whose view-dependent opacity falls below 0.01 at the target view during rendering. As shown in Tab.~\ref{tab:rendering_time}, this substantially reduces the number of active primitives.

\begin{table}[!h]
    \centering
    \resizebox{1.0\columnwidth}{!}{
        \begin{tabular}{lcccccc}
        \toprule
        \# views & PSNR diff. & SSIM diff. & LPIPS diff. & Avg. \# GS & Avg. pruned ratio & Render time (s)  \\
        \midrule
        32 & -0.02 & -0.01 &-0.01 & 3,392,471 &79.7\% & 0.003 \\ 
        64 & -0.06 & 0 &-0.01 & 5,347,737 &84.0\% & 0.005 \\     
        128 & -0.15 & 0 &0 & 8,556,380 &87.2\% & 0.008 \\   
        256 & -0.15 & 0 &-0.01 & 14,572,584 &89.1\% & 0.012 \\
        \bottomrule
        \end{tabular}        
    }  
    \vspace{-1.5mm}
    \caption{Rendering analysis after redundancy pruning. Rendering times are measured on a RTX 4090 GPU and correspond to the time required to render a single image.}
    \vspace{-2mm}
    \label{tab:rendering_time}
\end{table}

\vspace{-6mm}
\section{Conclusion and discussions}
\label{sec:discussions}
\vspace{-2.5mm}
We introduce MVP, a highly scalable and efficient transformer architecture for multi-view reasoning. The core of our approach is a novel dual attention hierarchy on two complementary axes: an inter-view hierarchy that simultaneously expands its attention scope from local per-frame reasoning, through intermediate group-wise attention, to a full global context, and an intra-view hierarchy that progressively aggregates spatial features. By fusing features from all stages via a pyramidal feature aggregation module, our model produces high-fidelity results with remarkable inference speed, enabling high-quality multi-view 3D reconstruction from a large number of input views in a single pass. While our current experiments focus on the posed, feed-forward static 3D reconstruction, we believe MVP can be extended to dynamic scenes or visual geometry related tasks with minimal architectural 
changes.

\section*{Acknowledgements}
This research was supported by the Ministry of Science and ICT (MSIT) of Korea, under the National Research Foundation (NRF) grant (RS-2024-00337548); the Korea-US Collaborative Research Fund (KUCRF), funded by the Ministry of Science and ICT and Ministry of Health \& Welfare, Republic of Korea (RS-2024-00468417); Samsung Research Funding \& Incubation Center of Samsung Electronics under Project Number SRFC-IT2401-01; and the Artificial Intelligence Industrial Convergence Cluster Development Project funded by the Ministry of Science and ICT (MSIT, Korea) \& Gwangju Metropolitan City.
This work was also supported by the Institute of Information \& Communications Technology Planning \& Evaluation (IITP) grant funded by the Korea government (MSIT): No. RS-2024-00457882, AI Research Hub Project; No. RS-2025-25441838, Development of a Human Foundation Model for Human-Centric Universal Artificial Intelligence and Training of Personnel; and No. RS-2020-II201361, Artificial Intelligence Graduate School Program (Yonsei University).
{
    \small
    \bibliographystyle{ieeenat_fullname}
    \bibliography{main}
}

\appendix

\twocolumn[
    \centering
    \Large
    \textbf{\thetitle}\\
    \vspace{0.5em}Supplementary Material \\
    \vspace{1.0em}
] %

\section{Additional Details.} 
\noindent\textbf{Datasets.} We adopt the official  DL3DV~\cite{ling2024dl3dv} split for benchmark dataset. In total, we use 9,995 scenes for training and 140 scenes for benchmarking, with no overlap between the two splits. For zero-shot inference, we use the \texttt{train} and \texttt{truck} scenes in Tanks\&Temples~\cite{Knapitsch2017tnt}, following \cite{kerbl20233dgs, ziwen2025longlrm}, and 9 scenes in Mip-NeRF360~\cite{barron2022mipnerf360} dataset. Considering our evaluation resolution of 960$\times$540, we use downsampled images (from the original images) whose resolution is closest to, but not smaller than, this target resolution (Tab.~\ref{tab:mipnerf360_resolution}).

\begin{table}[!h]
    \centering
    \resizebox{0.75\linewidth}{!}{
        \begin{tabular}{cccc}
        \toprule
        scene name & downsample & scene name & downsample \\
        \midrule
        \texttt{bicycle} & 4 & \texttt{room} & 2 \\
        \texttt{bonsai} & 2 & \texttt{stump} & 4 \\
        \texttt{counter} & 2 & \texttt{flower} & 4 \\
        \texttt{garden} & 4 & \texttt{treehill} & 4 \\
        \texttt{kitchen} & 2 & & \\
        \bottomrule
        \end{tabular}
    }
     \vspace{-2mm}
    \caption{Resolution of Mip-NeRF360 evaluation dataset.}        
    \label{tab:mipnerf360_resolution}             \vspace{-2mm}
\end{table}

During evaluation, we select every eighth frame in each sequence as target views:
\begin{itemize}
    \item \textbf{DL3DV}: For input views, we follow Long-LRM~\cite{ziwen2025longlrm} and adopt their released indices for 16 to 128 input images. For additional settings with 192 and 256 input images, we uniformly sample input views from the remaining frames after excluding all target indices.
    \item \textbf{Tanks\&Temples}: For the 32-view setting, we use the input-view indices provided by Long-LRM, while for the 64- and 128-view settings, input views are uniformly sampled from non-target frames.
    \item \textbf{Mip-NeRF360}: For all view configurations (32, 64, 128), input views are uniformly sampled from the frames excluding the target indices.
\end{itemize}
Importantly, target view indices are kept fixed across all input-view configurations for each scene.

\noindent\textbf{Implementation details.} The proposed MVP architecture consists of 14 transformer blocks. Within each attention~\cite{vaswani2017attention} block, we first apply LayerNorm~\cite{ba2016layernorm} to the input, and then perform QK-Norm~\cite{henry2020qknorm} on the query and key projections using an RMSNorm~\cite{zhang2019rmsnorm} layer. Each block comprises a multi head attention module with head dimension 64, followed by a two layer feed forward network with GELU~\cite{hendrycks2016gelu} activation. For the spherical harmonics representation, we set the degree to 1 for color and 2 for opacity.

For the post-activation parameterization of 3D Gaussians~\cite{kerbl20233dgs}, we follow the implementations from Long-LRM and iLRM~\cite{kang2025ilrm}. For rendering, we employ the gsplat~\cite{ye2025gsplat} library for efficient differentiable rasterization.

We employ PRoPE~\cite{li2025prope}, which represents camera poses as relative positional signals within the attention mechanism. Unlike the original implementation, we use a ``Plücker (extrinsics and intrinsics) rays + PRoPE" formulation instead of ``Cam (intrinsics only) rays + PRoPE", as we found this variant to yield better empirical performance. 

\begin{figure}[!h]
   \centering
   \includegraphics[width=1.0\linewidth]{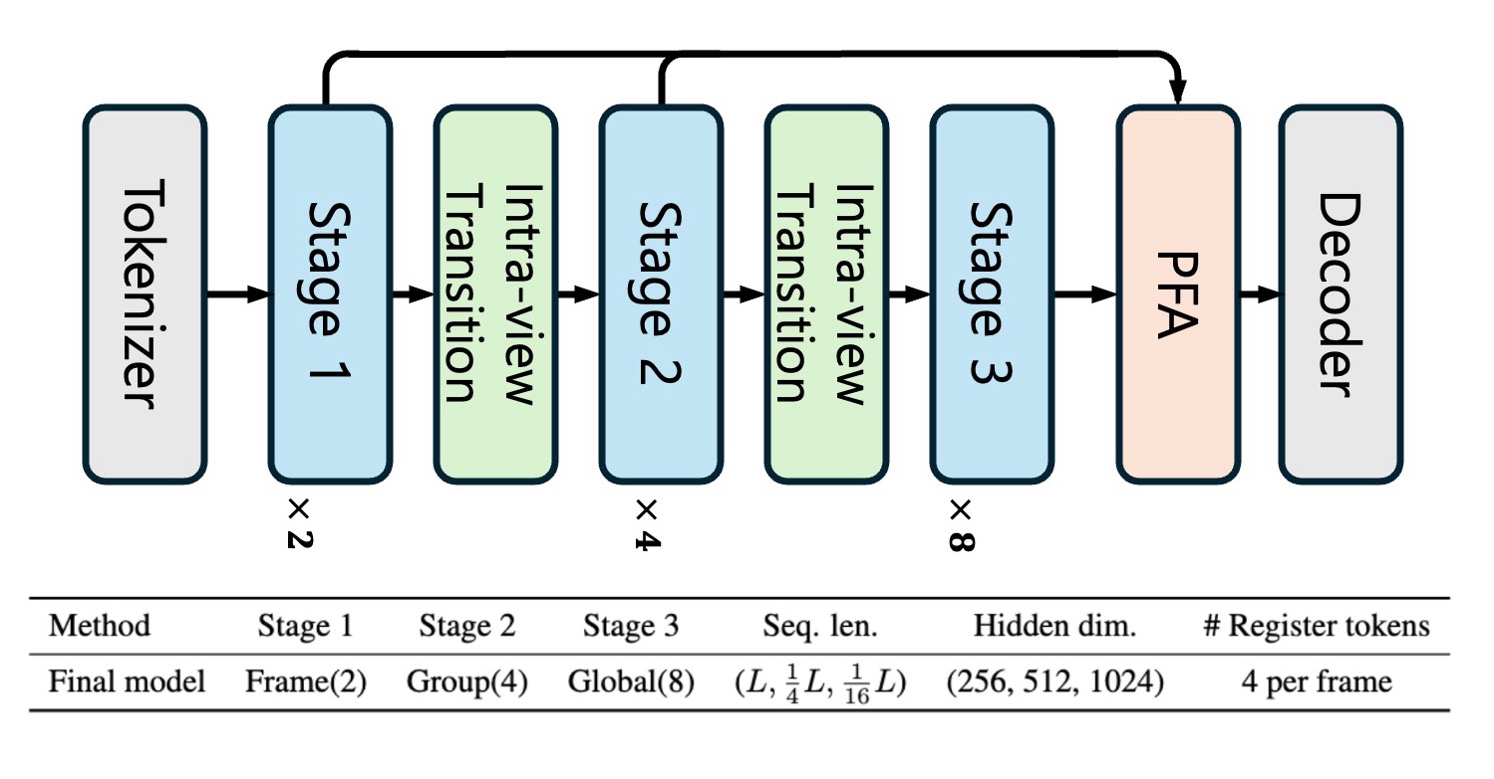}
   \caption{Architecture overview for our final MVP model. MVP employs a multi-stage hierarchy performing self-attention across increasing frame coverage: frame-wise, group-wise, and global. Each stage is linked by an Intra-view Transition layer using convolution for spatial downsampling and channel up-projection. Finally, a Pyramidal Feature Aggregation module fuses multi-level representations into a unified feature map.}
   \label{fig:architecture_overview}
   \vspace{-1.5mm}
\end{figure}

\begin{table*}[!t]
    \centering
    \resizebox{\linewidth}{!}{
        \setlength{\tabcolsep}{2pt}
        \renewcommand{\arraystretch}{1.0}
        \begin{tabular}{@{}lcccccccccccccccc@{}}
        \toprule
        \multirow{2}{*}{Method} & \multicolumn{4}{c}{16 views} & \multicolumn{4}{c}{32 views} & \multicolumn{4}{c}{64 views} & \multicolumn{4}{c}{128 views} \\
        \cmidrule(lr){2-5}\cmidrule(lr){6-9}\cmidrule(lr){10-13}\cmidrule(lr){14-17}          & PSNR $\uparrow$ & SSIM $\uparrow$ & LPIPS $\downarrow$ & Time (s) $\downarrow$ & PSNR $\uparrow$ & SSIM $\uparrow$ & LPIPS $\downarrow$ & Time (s) $\downarrow$ & PSNR $\uparrow$ & SSIM $\uparrow$ & LPIPS $\downarrow$ & Time (s) $\downarrow$ & PSNR $\uparrow$ & SSIM $\uparrow$ & LPIPS $\downarrow$ & Time (s) $\downarrow$ \\ 
         \midrule
        3D-GS$_{30k}$~\cite{kerbl20233dgs} & 21.96 & 0.766 & 0.237 & 8min & 25.09 & 0.838 & 0.175 & 8min & 28.02 & 0.890 & 0.134 & 8min & 29.88 & 0.917 & 0.115 & 8min \\
        \midrule
        Long-LRM~\cite{ziwen2025longlrm} & 20.65 & 0.707 & 0.328 & 0.50 & 23.54 & 0.776 & 0.270 & 0.84 & 23.15 & 0.787 & 0.263 & 2.08 & 20.78 & 0.741 & 0.307 & 6.39 \\
        iLRM~\cite{kang2025ilrm} & 21.62 & 0.746 & 0.316 & 0.19 & 23.93 & 0.800 & 0.259 & 0.53 & 24.11 & 0.816 & 0.243 & 1.66 & 22.72 & 0.804 & 0.251 & 5.61 \\
        Ours & \best{23.49} & \best{0.799} & \best{0.238} & \best{0.09} & \best{25.75} & \best{0.848} & \best{0.186} & \best{0.17} & \best{27.62} & \best{0.883} & \best{0.154} & \best{0.36} & \best{28.80} & \best{0.903} & \best{0.136} & \best{0.77} \\
        \bottomrule
        \end{tabular}
    }
    \vspace{-2mm}
    \caption{
    Quantitative comparisons on the DL3DV evaluation dataset with varying numbers of input views. For all metrics, we report the results by re-evaluating the models from their 32-view checkpoints.}      
    \label{tab:quantitative result on dl3dv}   \vspace{-2mm} 
\end{table*}   

\begin{table}[!t]
    \centering
    \resizebox{1.0\columnwidth}{!}{
        \setlength{\tabcolsep}{2pt}
        \renewcommand{\arraystretch}{1.1}
        \begin{tabular}{@{}lcccccccc@{}}
        \toprule        
        \multirow{2}{*}{Method} & \multicolumn{4}{c}{192 views} & \multicolumn{4}{c}{256 views} \\
        \cmidrule(lr){2-5}\cmidrule(lr){6-9} & PSNR $\uparrow$ & SSIM $\uparrow$ & LPIPS $\downarrow$ & Time (s) $\downarrow$ & PSNR $\uparrow$ & SSIM $\uparrow$ & LPIPS $\downarrow$ & Time (s) $\downarrow$ \\
        \midrule
        3D-GS$_{30k}$~\cite{kerbl20233dgs} & 30.53 & 0.926 & 0.109 & 8min & 30.75 & 0.929 & 0.107 & 8min \\
        \midrule
        Long-LRM~\cite{ziwen2025longlrm} & \multicolumn{8}{c}{OOM (GPU memory limit exceeded, 80\,GB)} \\
        iLRM~\cite{kang2025ilrm} & 21.57 & 0.787 & 0.267 & 11.91 & 20.62 & 0.768 & 0.283 & 20.92 \\
        Ours & \best{29.32} & \best{0.911} & \best{0.130} & \best{1.23} & \best{29.42} & \best{0.913} & \best{0.128} & \best{1.84} \\
        \bottomrule        
        \end{tabular}
    }   
    \vspace{-2mm}
    \caption{Quantitative comparisons on the DL3DV evaluation dataset with varying numbers of input views.}   
    \label{tab:quantitative result on dl3dv2}      
    \vspace{-2mm}
\end{table}

\noindent\textbf{Training details.} We train MVP transformer using a three-stage training schedule:
\begin{itemize}
    \item \textbf{First stage}: We train at 480$\times$256 resolution for 100k iterations with a learning rate of $2e^{-4}$. Each training sample uses 32 input views and 12 target views, with a batch size of 8 per GPU, resulting in a total batch size of 256. The frame interval is uniformly sampled between 64 and 128.
    \item \textbf{Second stage}: We increase the resolution to 960$\times$540 and train for 50k iterations with a learning rate of $2e^{-5}$. In this stage, we use 32 input views and 6 target views, with a batch size of 2 per GPU for a total batch size of 64. We sample input views from the full frame range and apply the intrinsic augmentation strategy proposed in~\cite{ziwen2025longlrm}.
    \item \textbf{Third stage}: We keep the resolution at 960$\times$540 and further train for 30k iterations with a learning rate of $2e^{-5}$, using various numbers of input and target views to improve robustness across different view configurations. We again draw input views from the entire frame range, while continuing to use the intrinsic augmentation strategy.
\end{itemize}
For all stages, we use a cosine learning rate scheduler (with a 3k-step warmup in the first stage only) and the AdamW~\cite{loshchilov2017adamw} optimizer with $\beta_1=0.9$ and $\beta_2=0.95$, and a weight decay of $0.05$. Weight decay is not applied to normalization and bias parameters. We also employ EMA~\cite{ema} and do not use gradient clipping, following \cite{nair2025scalingLVSM}. To improve training efficiency, we employ FlashAttention2~\cite{dao2023flash-v2} and gradient checkpointing~\cite{chen2016checkpointing}. We also use mixed-precision training with bfloat16 to speed up optimization while preserving numerical stability.

Tab.~\ref{tab:various_input} provides our training configurations with different numbers of input views. We adjust the batch size and the number of target renderings to balance iteration time and memory consumption. As described in the main manuscript, we freeze the frame- and group-wise attention blocks and train only the global attention blocks, which enables efficient fine-tuning while preserving the learned local and group representations.

\begin{table}[!h]
    \centering
    \resizebox{0.6\linewidth}{!}{
        \begin{tabular}{ccccc}
        \toprule
        \# Input views & 16 & 32 & 64 & 128 \\
        \midrule
        Batch size (per GPU) & 4 & 2 & 1 & 1 \\
        \# Target views & 6 & 6 & 6 & 1 \\
        \bottomrule
        \end{tabular}
    }
    \vspace{-2mm}
    \caption{Training configurations for varying input views.}
    \label{tab:various_input}  
    \vspace{-2mm}
\end{table}

\section{Additional Results}
We additionally evaluate our method on the recently released DL3DV~\cite{ling2024dl3dv} evaluation split, which comprises 51 scenes in our experiments. Tab.~\ref{tab:quantitative result on dl3dv} and~\ref{tab:quantitative result on dl3dv2} present quantitative results comparing our approach with 3D Gaussian Splatting~\cite{kerbl20233dgs} (3D-GS), Long-LRM~\cite{ziwen2025longlrm}, and iLRM~\cite{kang2025ilrm}. Our method surpasses existing feed-forward approaches by a large margin in both reconstruction quality and inference efficiency. We evaluate 16-view metrics of Long-LRM and iLRM using checkpoints trained with 32 input views.

\begin{figure}[!h]
   \centering
   \includegraphics[width=1.0\linewidth]{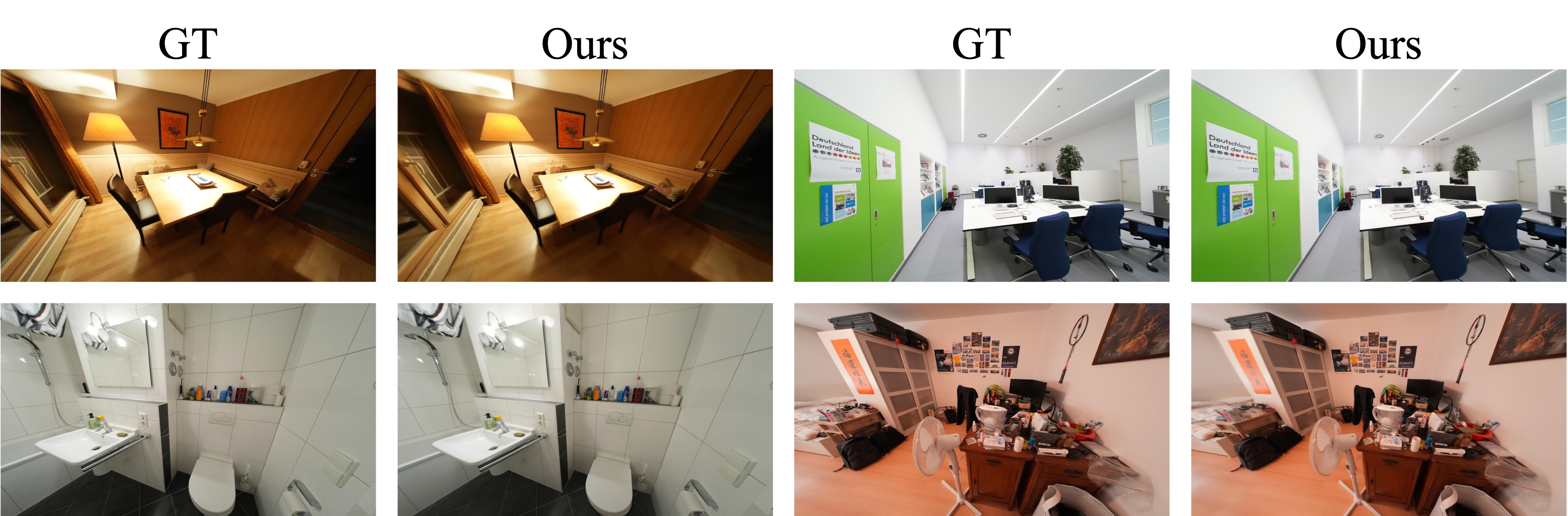}
   \vspace{-5mm}
    \caption{Qualitative performance on the ScanNet++ dataset.}   
   \label{fig:scannet_sup}
   \vspace{-1.5mm}
\end{figure}

\noindent\textbf{Additional zero-shot comparisons} We also evaluate our method on the real-world ScanNet++~\cite{yeshwanthliu2023scannetpp} dataset in Tab.~\ref{tab:quantitative result on scannetpp} under sparse input-view settings, which naturally reflects robustness to larger viewpoint variations (Fig.~\ref{fig:scannet_sup}). 

\begin{table}[!h]
    \centering
    \resizebox{1.0\columnwidth}{!}{
        \begin{tabular}{lccccccccc}
        \toprule
        \multirow{2}{*}{Method} & \multicolumn{3}{c}{16 views} & \multicolumn{3}{c}{32 views} & \multicolumn{3}{c}{64 views} \\
        \cmidrule(lr){2-4}\cmidrule(lr){5-7}\cmidrule(lr){8-10}
        & PSNR & SSIM & LPIPS & PSNR & SSIM & LPIPS & PSNR & SSIM & LPIPS\\
        \midrule
        Long-LRM & 24.42 & 0.873 & 0.224 & 28.31 & 0.909 & 0.195 & 27.20 & 0.905 & 0.207\\     
        iLRM & 26.78 & 0.896 & 0.240 & 28.45 & 0.913 & 0.214 & 28.07 & 0.914 & 0.213\\   
        Ours & \best{27.41} & \best{0.900} & \best{0.198} & \best{28.63} & \best{0.910} & \best{0.187} & \best{29.11}  & \best{0.916} & \best{0.185} \\           
        \bottomrule
        \end{tabular}        
    }
    \vspace{-2mm}
    \caption{Quantitative comparisons on the ScanNet++ dataset under sparse input view setting.}
    \label{tab:quantitative result on scannetpp}
\end{table}

\begin{figure*}[!h]
    \centering
    \includegraphics[width=1.0\textwidth]{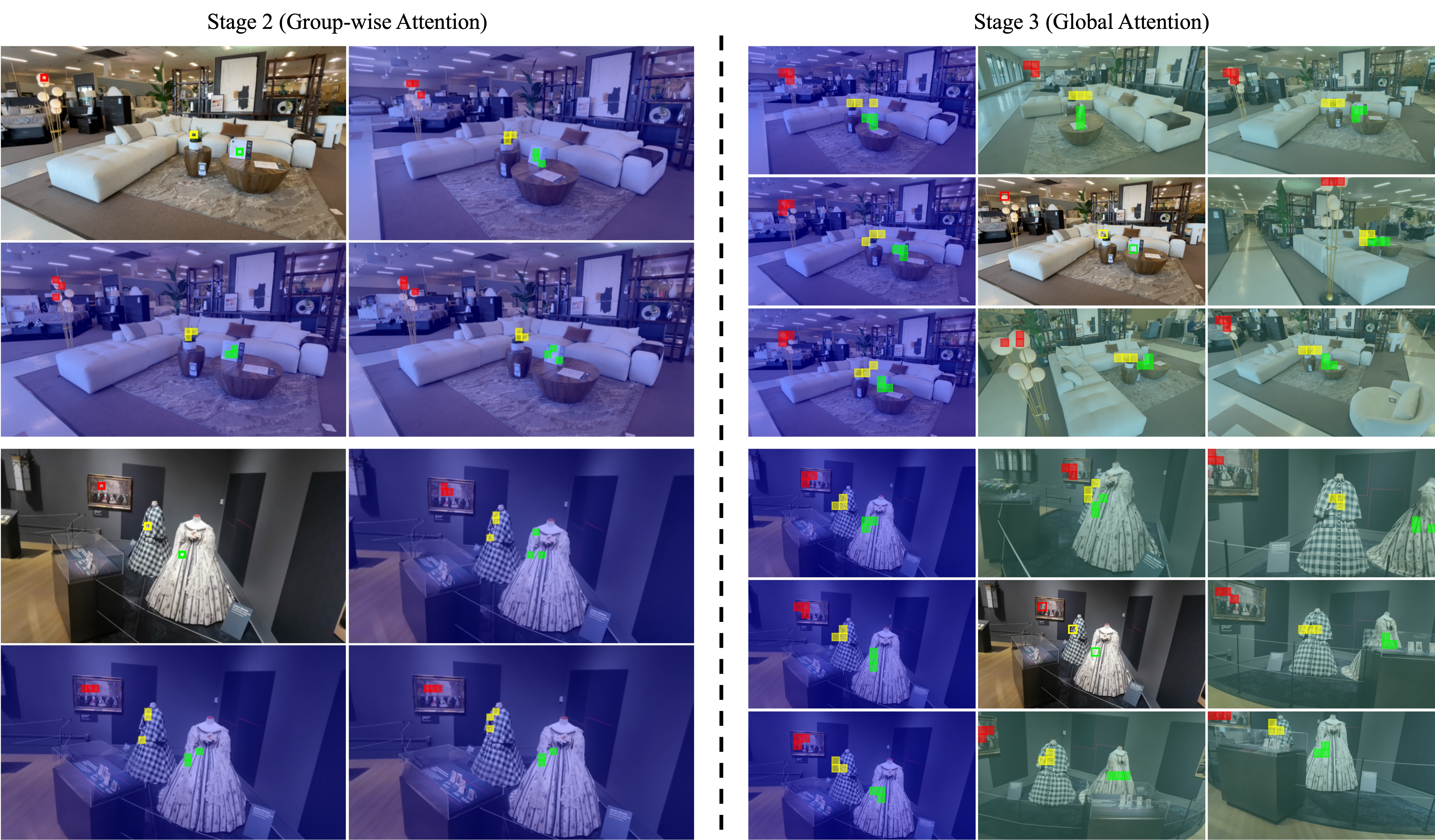} 
    \vspace{-6mm}    
    \caption{Attention visualization. For colored query patches (\textcolor[HTML]{C02A1E}{red}, \textcolor[HTML]{D1D147}{yellow}, \textcolor[HTML]{62CF42}{green}) in the reference view, we highlight top-3 attended tokens: on the left, tokens attended within the group (blue overlay), and on the right, tokens attended within and outside the group (green overlay).}
    \vspace{-1mm}
    \label{fig:att_sub}
\end{figure*}

\noindent\textbf{Additional attention visualization.} Using the first frame as the reference view, we begin by selecting three query patches. For each query patch, we visualize its top-3 attended tokens from other viewpoints across different stages. At stage 2, we display the attended tokens restricted to the same group, whereas at stage 3, we additionally include tokens attended from views outside the group (Fig.~\ref{fig:att_sub}).

\noindent\textbf{Additional design choice.} We ablate the number of views per group (2, 4, and 8) in Tab.~\ref{tab:groupsize}. In addition to our primary objective of novel view synthesis, we also evaluate the model on a spatial cognition task~\cite{li2025prope}. In this task, one camera pose in a set of image–camera pairs is corrupted and the model must identify the mismatched pair, which probes its multi-view awareness. We replace the 3D-GS linear decoder with a head that produces a single scalar per token. For each input view, these scalars are averaged across all tokens to yield a single score, and a softmax over all views then produces a probability vector indicating which image–camera pair is most likely to be corrupted. A group size of 4 (baseline) offers consistently strong accuracy across different numbers of views, while groups of 2 provide too little context and groups of 8 bring only marginal gains with higher computational cost. Therefore, we adopt a group size of four as our default setting.

\begin{table}[!h]
    \centering
    \resizebox{1.0\columnwidth}{!}{
        \begin{tabular}{lcccccc}
        \toprule
        \multirow{2}{*}{Method} & \multicolumn{3}{c}{Novel view synthesis} & \multicolumn{3}{c}{Spatial cognition} \\
        \cmidrule(lr){2-4}\cmidrule(lr){5-7}     
        & PSNR $\uparrow$ & SSIM $\uparrow$ & LPIPS $\downarrow$ & 8 views & 16 views & 32 views \\
        \midrule
        Group 2  & 22.94 & 0.711 & 0.299 & 51.3\% & 75.7\% & 83.6\% \\
        Group 4 (baseline) & 23.18 & 0.716 & 0.300 & 77.1\% & 91.4\% & 96.4\% \\
        Group 8 & 23.18 & 0.717 & 0.297 & 84.3\% & 90.7\% & 97.1\% \\
        \bottomrule
        \end{tabular}        
    }   
    \caption{Ablation studies on the number of views per group.}
    \label{tab:groupsize} 
    \vspace{-3mm}
\end{table}

We also ablate the usage of register tokens in Tab. ~\ref{tab:Register}. Register token~\cite{darcet2023vision} was introduced to mitigate abnormally high token norm values and artifacts in the attention maps of large Vision Transformer models, which often lead to attention artifacts and degraded dense prediction performance. Consistent with this observation, we find that the variant without register tokens exhibits higher $L2$ norms across frames, while incorporating register tokens yields modest performance gains. The configuration for this ablation is the same as that of the ablation studies in the main manuscript.

\begin{table}[!h]
    \centering  
    \resizebox{1.0\columnwidth}{!}{
        \begin{tabular}{lcccccccc}
        \toprule
        \multirow{2}{*}{Method} & \multirow{2}{*}{PSNR} & \multirow{2}{*}{SSIM} & \multirow{2}{*}{LPIPS} & \multicolumn{3}{c}{Avg. intra-frame feature norm} \\
        \cmidrule(lr){5-7}
        & & & & Stage 1 & Stage 2 & Stage 3\\
        \midrule
        Baseline & 22.79 & 0.733 & 0.235 & 14.01 & 84.49 & 113.36\\     
        w/o Register & 22.52 & 0.723 & 0.242 & 15.98 & 286.61 & 229.82\\     
        \bottomrule
        \end{tabular}        
    }
    \vspace{-1mm}
    \caption{Ablation study on register tokens.}
    \label{tab:Register}
    \vspace{-2mm}
\end{table}

\noindent\textbf{Point map estimation.} In addition to novel view synthesis, we also evaluate our method on point map estimation, which quantifies the geometric accuracy of the reconstructed 3D scenes represented with explicit primitives. We evaluate point map accuracy on NRGBD~\cite{azinovic2022nrgbd} and ETH3D~\cite{eth}, and report the Chamfer distance and the F1-score. Specifically, we first back-project the ground-truth depth maps into 3D point clouds using the camera poses. We then rigidly align the predicted point clouds to these reference points with the Umeyama algorithm~\cite{umeyama}, and finally apply the masks to discard points in invalid regions. We use an image resolution of 960$\times$540, which matches the training resolution for both our method and the baselines. It is important to note that both our approach and the baselines are trained using only a photometric rendering loss from 3D-GS, in contrast to geometry-supervised methods that exploit ground-truth depth or point clouds~\cite{wang2024dust3r, wang2025vggt, Yang_2025_Fast3R, wang2025pi3}. Consequently, the numbers in Tab.~\ref{tab:pointmap} should be read primarily as evidence of relative improvements within this setting, rather than as a direct comparison to fully geometry-supervised models. Overall, our method outperforms the baseline, even though the baseline is additionally regularized using pretrained DepthAnything~\cite{depthanything} model during training.

\begin{table}[!h]
    \centering
    \resizebox{0.9\columnwidth}{!}{
        \begin{tabular}{lccccc}
        \toprule
        \multirow{2}{*}{Method} & \multirow{2}{*}{Views} & \multicolumn{2}{c}{NRGBD~\cite{azinovic2022nrgbd}} & \multicolumn{2}{c}{ETH3D~\cite{eth}} \\
        \cmidrule(lr){3-4}\cmidrule(lr){5-6}     
        & & CD $\downarrow$ & F1-score $\uparrow$ & CD $\downarrow$ & F1-score $\uparrow$ \\
        \midrule
        Long-LRM~\cite{ziwen2025longlrm}  & \multirow{2}{*}{16} & 0.53 & 0.52 & 2.75 & 0.32 \\
        Ours & & \best{0.18} & \best{0.54} & \best{1.74} & \best{0.34} \\
        \midrule
        Long-LRM  & \multirow{2}{*}{32} & 0.43 & \best{0.59} & 2.69 & 0.39 \\
        Ours & & \best{0.14} & 0.56 & \best{2.22} & \best{0.42} \\        
        \bottomrule
        \end{tabular}        
    }
    \caption{Quantitative comparison of point map estimation. We set the threshold value for f1-score to 0.1. We exclude iLRM~\cite{kang2025ilrm} from this evaluation, as it predicts low-resolution Gaussians that would require additional upsampling, potentially degrading its performance and hurt a fair comparison.}
    \label{tab:pointmap}
    \vspace{-2mm}
\end{table}

\begin{figure}[!h]
    \centering
\includegraphics[width=1.0\columnwidth]{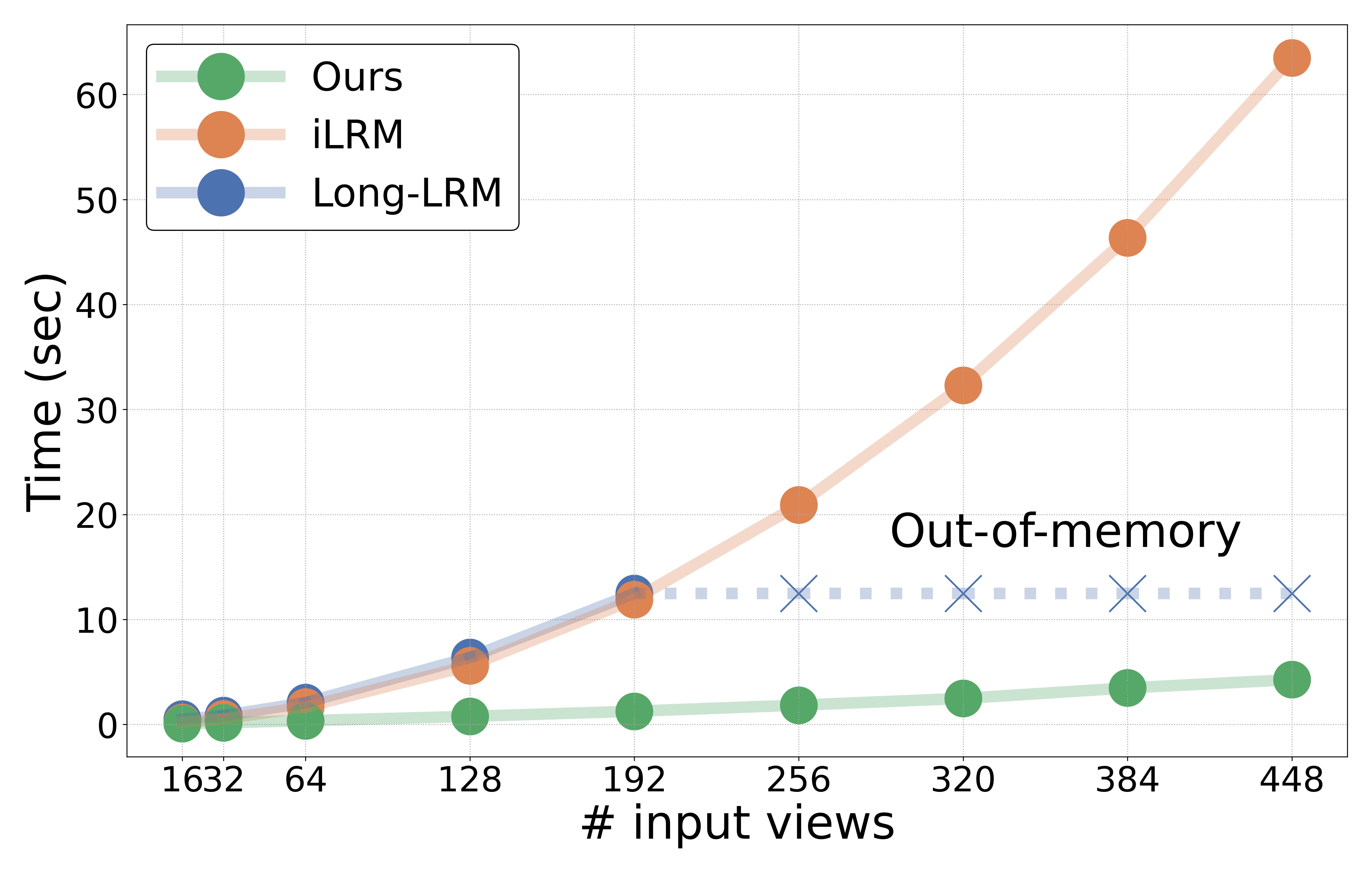}  
    \caption{Inference time comparison.}
    \label{fig:inference_time}
\end{figure}

\noindent\textbf{Inference time comparison.} Fig.~\ref{fig:inference_time} reports inference time as a function of the number of input views. We measure all timings at an input resolution of 960$\times$540 using the official implementations released by the respective authors.
Our method exhibits substantially better scalability than existing feed-forward baselines, achieving consistently lower latency across all view counts. Long-LRM encounters an out-of-memory issue on a 80GB GPU once the number of input views exceeds 192. Note that the reported inference time only accounts for the generation of 3D Gaussians. For novel view rendering, Long-LRM encounters a memory error when using more than 128 input views.

\noindent\textbf{FLOPs comparison.}
We report theoretical FLOPs comparisons or our model with global- and alternating-attention variants in Tab. ~\ref{tab:FLOPs}, assuming 14 transformer layers and identical input tokenizers. Group size for our model is set to 4. V, L, and D denote the number of views, the number of tokens per input view, and the hidden dimension, respectively.

\begin{table}[!h]
    \centering
    \resizebox{1.0\columnwidth}{!}{
        \begin{tabular}{lcccccc}
        \toprule
         \multirow{2}{*}{Method} & \multirow{2}{*}{Theoretical FLOPs} & \multicolumn{3}{c}{PFLOPs with (V, L, D) input}\\
        \cmidrule(lr){3-5}
        & & (32, 1920, 1024) & (32, 8160, 1024)& (128, 8160, 1024) \\
        \midrule
        Global. Att. & $VLD(336D + 56VL)$ & 0.24 & 4.0 & 62.93 \\
        Alter. Att. & $VLD(336D + 28(1+V)L)$ & 0.13 & 2.11 & 31.89 \\
        Ours & $VLD(21D+(3.3125+\frac{V}{16})L + 1.5)$ & 0.01 & 0.1 & 0.81\\
        \bottomrule
        \end{tabular}        
    }
    \caption{Theoretical FLOPs comparison.}
    \label{tab:FLOPs}
\end{table}

\noindent\textbf{Robustness to camera pose.} We evaluate the sensitivity of our method to inaccurate camera poses on DL3DV benchmark dataset (32-view, 960$\times$540) by adding random Gaussian noise with varying standard deviations to the rotation and translation components. As shown in Tab.~\ref{tab:noise}, performance degrades noticeably as noise increases, particularly for translation perturbations. This indicates that our method relies on reasonably accurate camera poses, which remains a limitation to be addressed in future work.

\begin{table}[!h]
    \centering
    \resizebox{1.0\columnwidth}{!}{
        \begin{tabular}{lccccccccc}
        \toprule
        \multirow{2}{*}{std.} & \multicolumn{3}{c}{0} & \multicolumn{3}{c}{0.001} & \multicolumn{3}{c}{0.005} \\
        \cmidrule(lr){2-4}\cmidrule(lr){5-7}\cmidrule(lr){8-10}
        & PSNR & SSIM & LPIPS & PSNR & SSIM & LPIPS & PSNR & SSIM & LPIPS\\
        \midrule   
        Rotation & 25.96 & 0.847 & 0.187 & 24.65 & 0.839 & 0.190 & 23.34  & 0.746 & 0.237 \\           
        Translation & - & - & - & 24.75 & 0.804 & 0.204 & 20.85  & 0.645 & 0.297 \\     
        \bottomrule
        \end{tabular}        
    }

    \caption{Robustness evaluations on camera pose.}
    \label{tab:noise}
\end{table}

\noindent\textbf{Remarks on posed setting.} We agree that known camera poses may limit applicability in some cases. Nevertheless, posed settings still often remain highly relevant in practice, including production-ready volumetric multi-view video systems, autonomous driving with calibrated cameras, and robotics applications where reasonably accurate poses can be estimated from inertial sensors. We also view posed multi-view modeling as an important step toward unposed settings, as evidenced by recent transformer-based approaches (e.g., GS-LRM~\cite{zhang2024gslrm}, LVSM~\cite{jin2025lvsm}) and their follow-ups (e.g., VGGT~\cite{wang2025vggt}, Depth Anything 3~\cite{lin2025depth}), which adopt multi-view transformers as a core component. We hope our work aligns with this evolution and provides a solid foundation for future pose-free extensions.

\begin{figure*}[!h]
    \centering
    \includegraphics[width=1.0\textwidth]{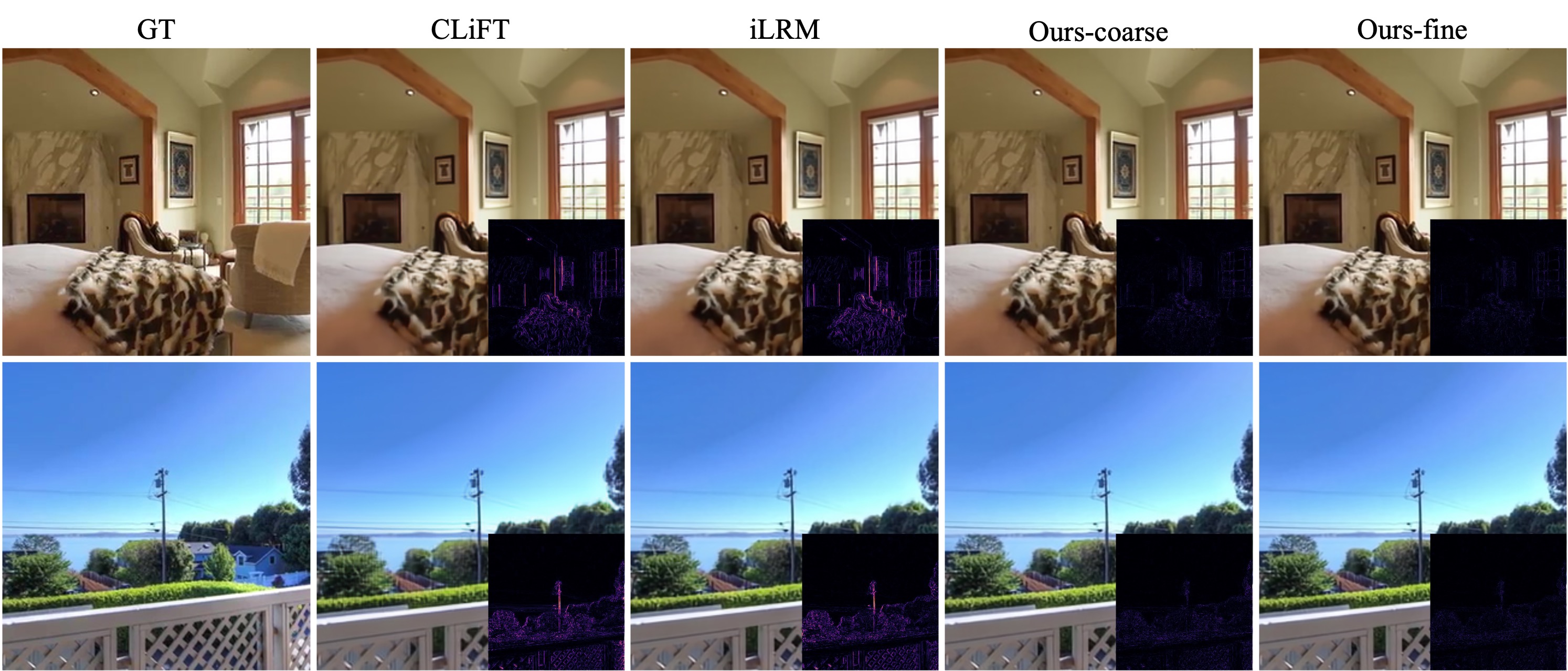} 
    \vspace{-6mm}    
    \caption{Qualitative results on the 4-view RE10K dataset. Per-pixel error maps are shown in the bottom-right corner of each image.}  
    \vspace{-1mm}
    \label{fig:re10k_sup}
\end{figure*}

\begin{figure*}[!h]
    \centering
    \includegraphics[width=1.0\textwidth]{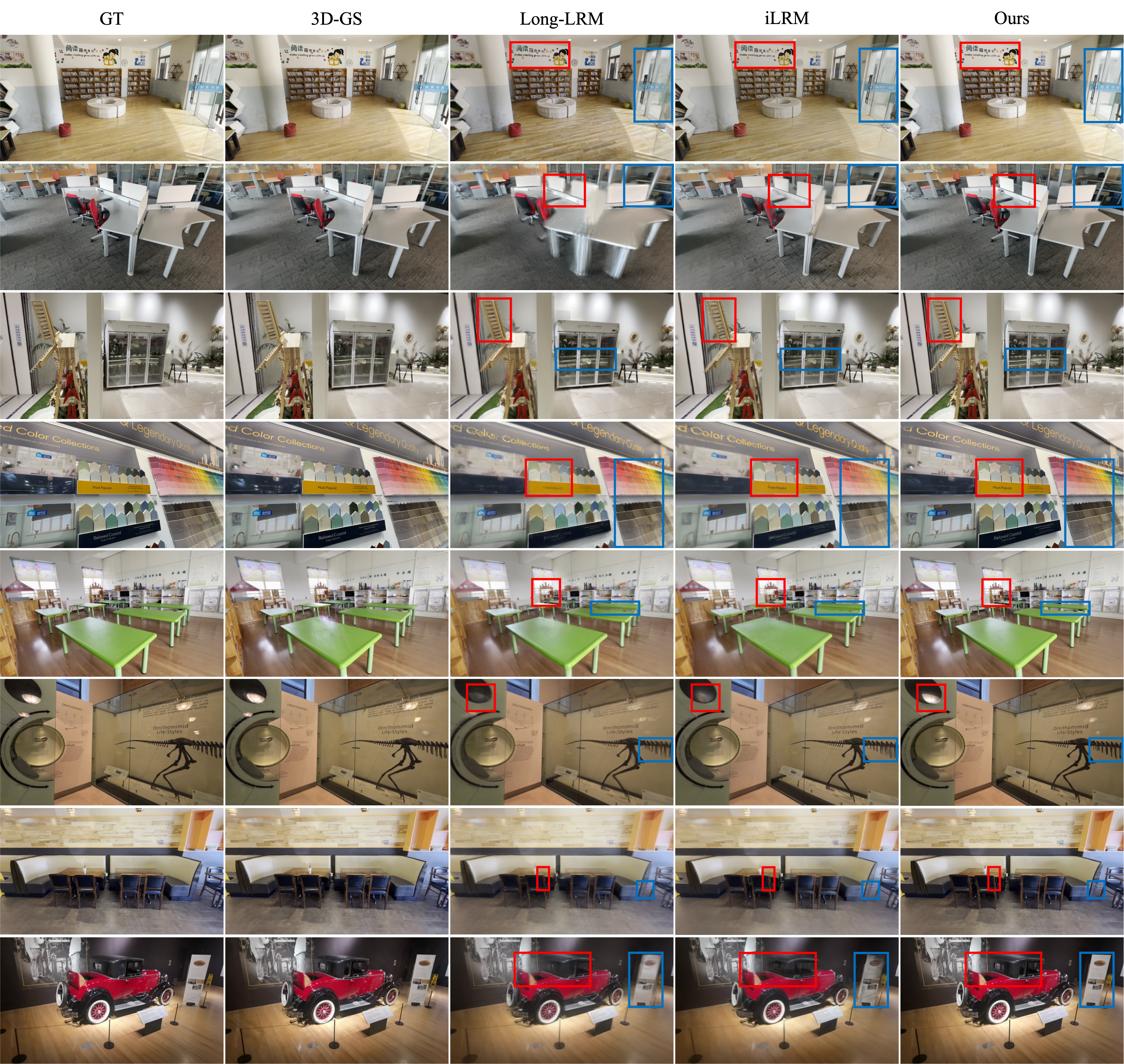} 
    \vspace{-6mm}    
    \caption{Qualitative results on the DL3DV-Benchmark across varying input view counts (128, 64, 32, and 16). The rows are arranged in descending order of view count, with two rows displayed for each setting.}  
    \vspace{-1mm}
    \label{fig:qual_sub1}
\end{figure*}

\begin{figure*}[!h]
    \centering
    \includegraphics[width=1.0\textwidth]{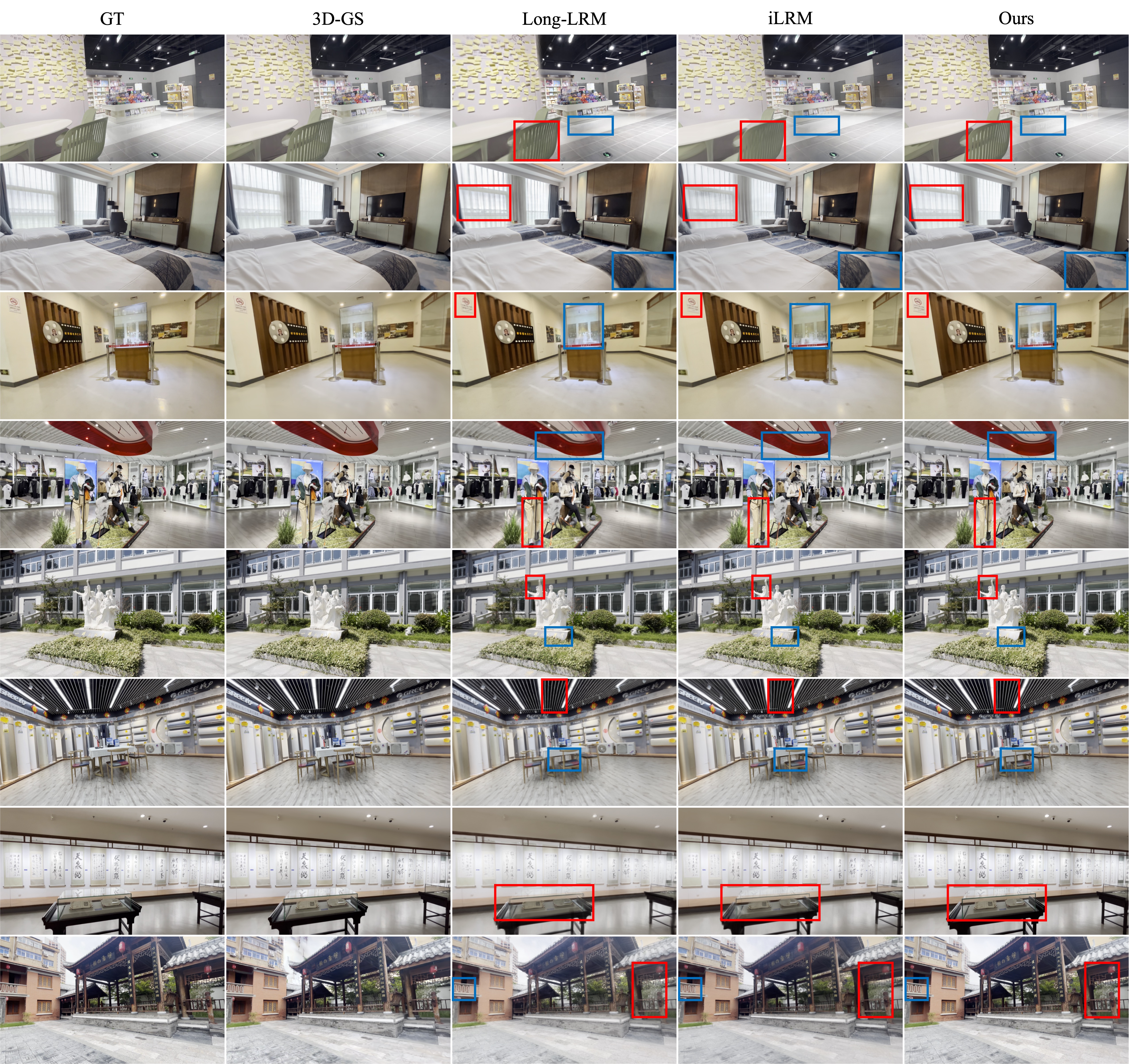} 
    \vspace{-6mm}    
    \caption{Qualitative results on the DL3DV-Evaluation across varying input view counts (128, 64, 32, and 16). The rows are arranged in descending order of view count, with two rows displayed for each setting.}  
    \vspace{-1mm}
    \label{fig:qual_sub2}
\end{figure*}

\begin{figure*}[!h]
    \centering
    \includegraphics[width=1.0\textwidth]{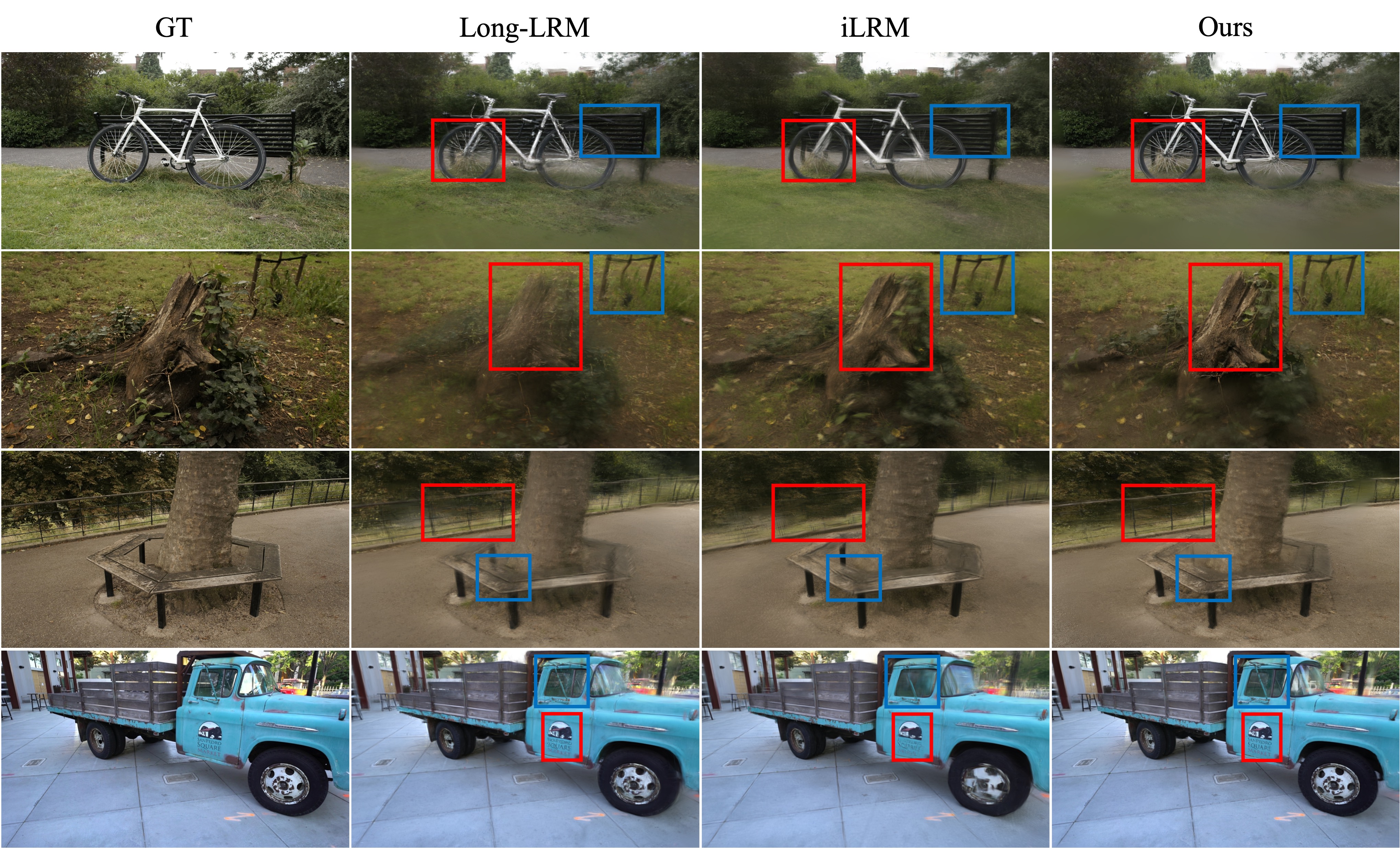} 
    \vspace{-5mm}    
    \caption{Qualitative results on the Mip-NeRF360 (top three rows), and \texttt{truck} scene from Tanks\&Temples (bottom row). We visualize our rendering results with 32 input views, showing that our method demonstrates clear improvements on generalization and multi-view consistency under sparse inputs. }  
    \vspace{-1mm}
    \label{fig:qual_sub3}
\end{figure*}

\begin{figure*}[!h]
    \centering
    \includegraphics[width=1.0\textwidth]{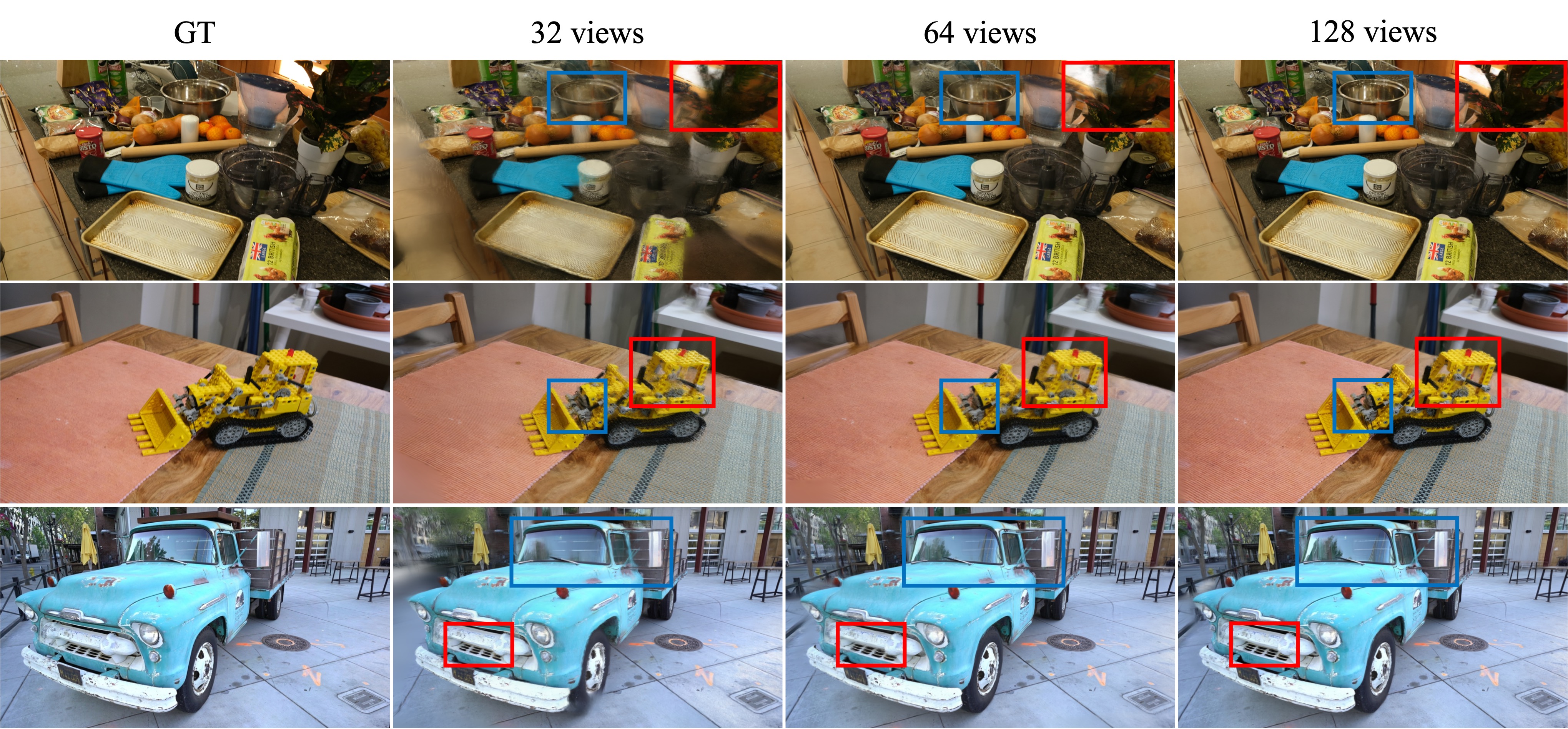} 
    \vspace{-6mm}    
    \caption{Qualitative results on the Mip-NeRF360 (top two rows), and \texttt{truck} scene from Tanks\&Temples (bottom row). We visualize our rendering results as the number of input views increases, revealing progressively improved image quality and demonstrating that our method scales effectively with the number of views.}  
    \vspace{-1mm}
    \label{fig:qual_sub4}
\end{figure*}

\begin{figure*}[!h]
    \centering
    \includegraphics[width=1.0\textwidth]{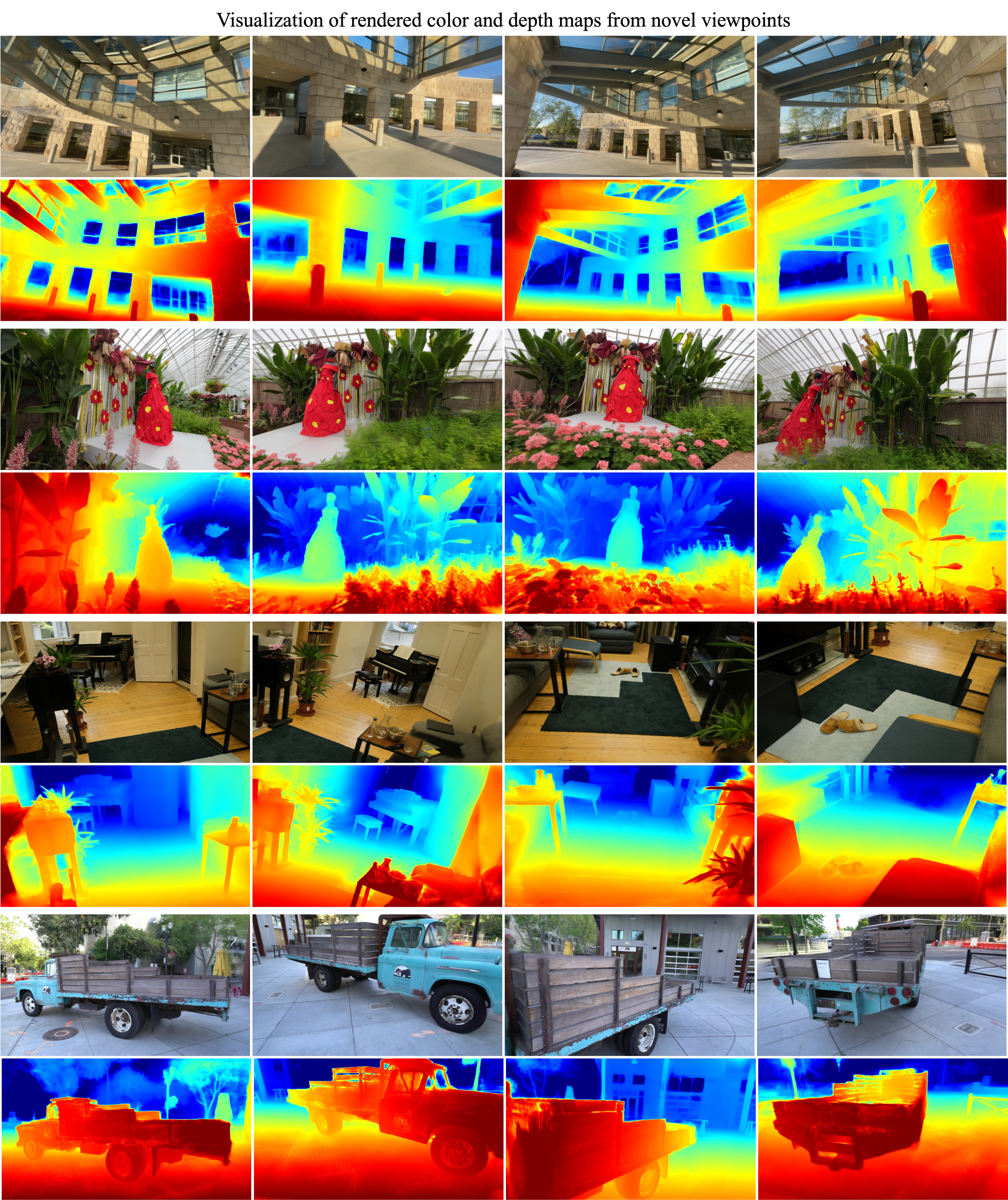} 
    \vspace{-6mm}    
    \caption{Qualitative visualization of rendered color and depth maps from novel viewpoints using 32 input images. Scenes from DL3DV (top four rows), Mip-NeRF360  (fifth and sixth row), and Tanks\&Temples (bottom two rows) are shown.}  
    \vspace{-1mm}
    \label{fig:qual_sub5}
\end{figure*}

\noindent\textbf{Additional qualitative results.} We provide further qualitative comparisons on the RE10K~\cite{zhou2018re10k}, DL3DV, Tanks\&Temples~\cite{Knapitsch2017tnt}, and Mip-NeRF360~\cite{barron2022mipnerf360} datasets in the remainder of this manuscript (Fig.~\ref{fig:re10k_sup},~\ref{fig:qual_sub1},~\ref{fig:qual_sub2},~\ref{fig:qual_sub3},~\ref{fig:qual_sub4} and~\ref{fig:qual_sub5}).

\clearpage

\clearpage

\end{document}